\documentclass[preprint]{elsarticle}

\usepackage{bm}
\usepackage{lineno,hyperref}
\usepackage{geometry}
\usepackage{amssymb}
\usepackage{mathtools}
\usepackage{enumitem}
\usepackage{bm}
\usepackage{xcolor}
\usepackage{tikz}
\usetikzlibrary{calc,trees,positioning,arrows,chains,shapes.geometric,%
decorations.pathreplacing,decorations.pathmorphing,shapes,%
matrix,shapes.symbols,plotmarks,decorations.markings,shadows}
\usetikzlibrary{decorations.pathreplacing}
\usetikzlibrary{chains, fit, positioning}
\usepackage{subcaption}
\usepackage{graphicx}  
\usepackage{stmaryrd}
\usepackage{multirow}

\newcommand{\kron}{\otimes}
\newcommand{\kr}{\odot}
\newcommand*\rfrac[2]{{}^{#1}\!/_{#2}}
\newenvironment{review}{\color{black}}{}
\newcommand{\rev}[1]{\begin{review}#1\end{review}}

\newenvironment{reviw}{\color{black}}{}
\newcommand{\revv}[1]{\begin{reviw}#1\end{reviw}}
 
 \newtheorem{thm}{Theorem}
 \newtheorem{lem}{Lemma}
 \newdefinition{rmk}{Remark}
  \newdefinition{con}{Conclusion}
 \newproof{pf}{Proof}
 \newproof{pot}{Proof of Theorem \ref{thm1}}

\modulolinenumbers[5]

\journal{Journal of Mechanical Systems and Signal Processing}

\begin{document}

\begin{frontmatter}

\title{Decoupling multivariate functions using a nonparametric filtered tensor decomposition}
\tnotetext[mytitlenote]{This work was supported by the Flemish fund for scientific research FWO under license number G0068.18N.}


\author[1]{Jan Decuyper\corref{mycorrespondingauthor}}
\cortext[mycorrespondingauthor]{Corresponding author}
\ead{jan.decuyper@vub.be}

\author[2]{Koen Tiels}
\author[2]{Siep Weiland}
\author[1]{Mark C.\ Runacres}
\author[1,2]{Johan Schoukens}

\address[1]{Vrije Universiteit Brussel, Pleinlaan 2, 1050, Brussels, Belgium}
\address[2]{Eindhoven University of Technology, Eindhoven, The Netherlands}

\begin{abstract}

Multivariate functions emerge naturally in a wide variety of data-driven models. Popular choices are expressions in the form of basis expansions or neural networks. While highly effective, the resulting functions tend to be hard to interpret, in part because of the large number of required parameters. Decoupling techniques aim at providing an alternative representation of the nonlinearity. The so-called decoupled form is often a more efficient parameterisation of the relationship while being highly structured, favouring interpretability. In this work two new algorithms, based on filtered tensor decompositions of first order derivative information are introduced. \rev{The method returns nonparametric estimates of smooth decoupled functions.} Direct applications are found in, i.a.\ the fields of nonlinear system identification and machine learning.
\end{abstract}


\end{frontmatter}


\section{Introduction}

Nonlinear function regression is an essential part of most nonlinear modelling techniques. Be it as an end product, or as an internal element of a dynamical model, a multivariate function is typically used to describe the nonlinear relationship amongst variables. In many applications, a data-driven approach is preferred, e.g.\ when the required knowledge to build a model from first-principles is unavailable, or when the model is only required to connect the dots, without revealing any potential insight into the relationship. Relying exclusively on data, also known as black-box modelling, entails that the required model-complexity is a priori unknown. Powerful regression techniques are therefore characterised by strong approximation properties. Popular choices are, for instance, neural networks, belonging to the class of universal approximators \cite{kratsios2021,cybenko1989}, and all sorts of basis expansions, given the resulting ease in parameter estimation when the basis functions are fixed \cite{hastie2001}. Inherent to both types is the tendency to result in large models, described by a large number of parameters. It is important to realise that distinct reasons lie at the origin of the bulky size of both types:
\begin{itemize}
\item \textbf{Networks} rely on width and/or depth to allow complexity to build. It is only by invoking the power of large numbers that seemingly simple neurons transcend their individual capabilities \cite{lu2017}.
\item \textbf{Basis expansions}, on the other hand, may feature a set of functions of which the complexity can vary. In this case, however, the basis remains fixed while there is no evidence that a sparse representation of the relationship can be obtained given the proposed basis. Related so-called dictionary methods \cite{brunton2016} rely on regularisation to enforce sparseness, neglecting the root problem, which is an unfavourable basis.    
\end{itemize}


These insights have led to the development of so-called decoupling techniques. Given a generic \rev{continuous real} multivariate nonlinear function\revv{\footnote{\revv{See section \ref{s:notation} for an overview of the notation used in this work.}}}
\begin{equation}
\label{e:1}
\boldsymbol{q} = \boldsymbol{f}(\boldsymbol{p}),
\end{equation}
with $\boldsymbol{q} \in \mathbb{R}^n$ and $\boldsymbol{p} \in \mathbb{R}^m$, the prime objective of a decoupling technique is to introduce an appropriate linear transformation of $\boldsymbol{p}$, denoted $\boldsymbol{V}$. The goal of this transformation is to retrieve an alternative basis in which the nonlinear relationship may be described by a minimal set of univariate \rev{continuous real} functions, also called \emph{branches}. What adds to the power of the new representation is that the functions, which make up the branches, are not fixed beforehand. An appropriate basis together with matching branch functions is the joint product of the decoupling procedure. The decoupled structure is then of the following form
\begin{equation}
\label{e:2}
\boldsymbol{f}(\boldsymbol{p}) = \boldsymbol{W} \boldsymbol{g}(\boldsymbol{V}^{\top} \boldsymbol{p}),
\end{equation}
where the $i$th function is $g_i(z_i)$ with $z_i = \boldsymbol{v}_i^{\top} \boldsymbol{p}$, emphasising that all functions are strictly univariate. A schematic representation of a classical multivariate function (hereafter referred to as coupled function) and a decoupled function are presented in the top section of Fig.~\ref{f:overview}. The decoupled form has a number of attractive features: relying exclusively on univariate functions, a very structured representation is obtained. Moreover, the one-dimensional functions are much more tractable, which may lead to insight into the relationship. Examples of structure retrieval in dynamical models may be found in \cite{decuyper2021,csurcsia2022}. Additionally, decoupled functions tend to be a much more efficient representation of the nonlinearity, resulting in parameter reduction and parsimonious models. Note that the decoupled structure may be seen as a network, with $\boldsymbol{V}$ acting as weights, while allowing tailored nonlinear branches, contrary to the predefined activation functions typically found in neurons. \rev{This entails that the decoupled representation inherits all favourable properties of network models, e.g.\ neural networks.}

The number of allowed univariate functions, denoted $r$, is crucial since it will determine whether the implied equivalence of \eqref{e:2} can be attained. A second linear transformation $\boldsymbol{W}$, maps the function back onto the outputs. The linear transformations then have the following dimensions: $\boldsymbol{V} \in \mathbb{R}^{m \times r}$ and $\boldsymbol{W} \in \mathbb{R}^{n \times r}$.

Decoupling is a generic tool which is of practical use in, amongst others, the fields of nonlinear system identification and machine learning. The technique is designed to post-process multivariate functions which emerge naturally in data-driven modelling tasks.
\subsection{Decoupling using first-order information}
\begin{figure}
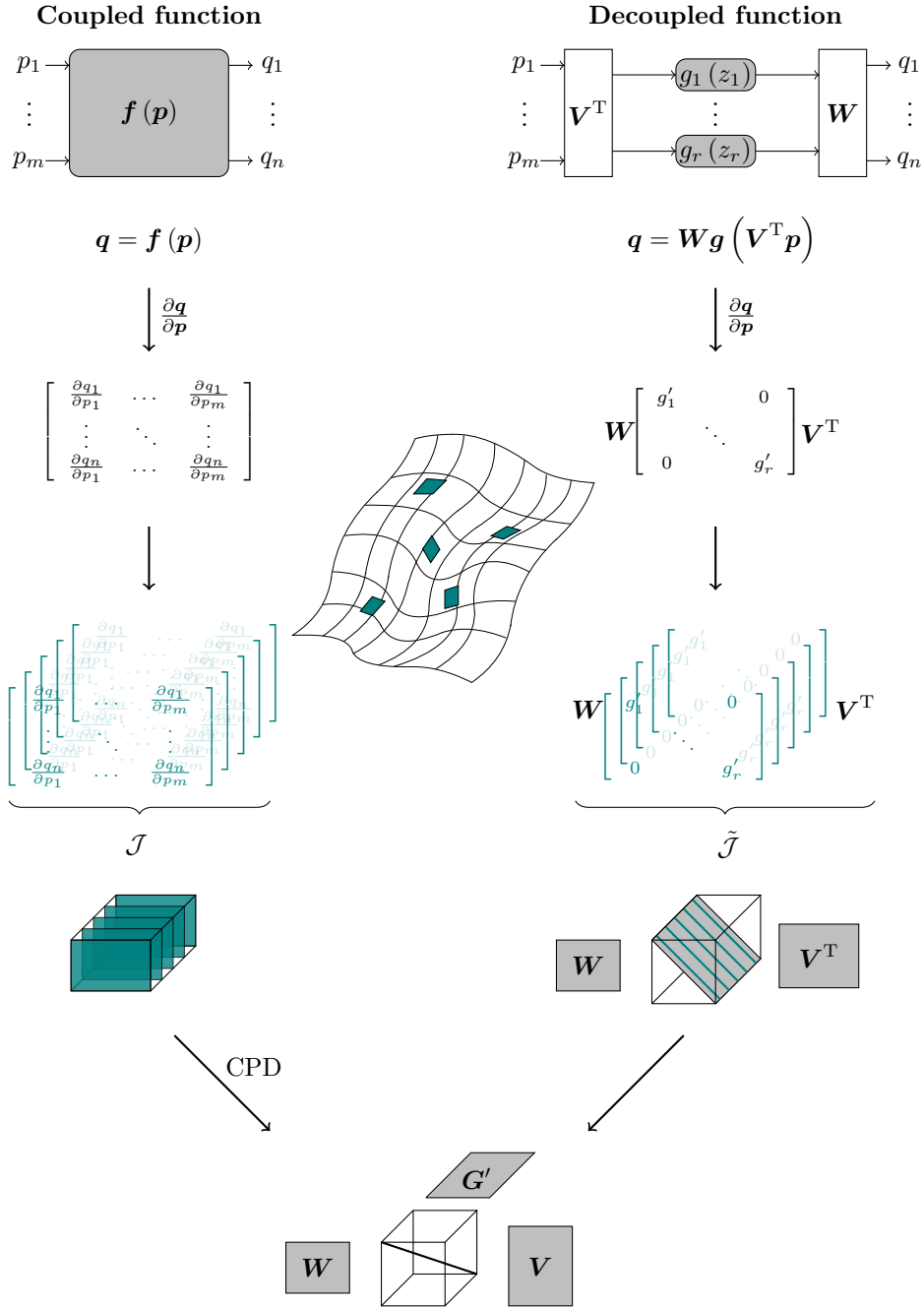

\begin{center}
\include{Figures/Jacobian_decomp_overview_tikz}
\vspace{-0.55cm}
\caption{Overview of the decoupling procedure of \cite{dreesen2014} based on first order derivative information. The column on the left illustrates the construction of a third order Jacobian tensor $\mathcal{J}$ out of evaluations of the Jacobian of a coupled function along a number of operating points. The column on the right illustrates the corresponding steps applied to a decoupled function. Both functions are linked by computing the CPD of $\mathcal{J}$, \color{black} i.e.\ an exact diagonal decomposition. If the CPD is unique, the approach retrieves an exact decoupled function. \color{black} Observe that independent of the nature and complexity of $\boldsymbol{q} = \boldsymbol{f}(\boldsymbol{p})$, a third order tensor decomposition problem is retrieved, whereas this is not the case for the alternative methods described in \cite{usevich2020}.}
\label{f:overview}
\end{center}
\end{figure}

A corollary of \eqref{e:2} is that the Jacobians of the left- and the right-hand side need to match. This was exploited by \cite{dreesen2014} when proposing a decoupling algorithm. Denoting \revv{an evaluation of} the Jacobian of the coupled function in point $\boldsymbol{p}(k)$ by $\boldsymbol{J}(k)$, 
\begin{equation}
\label{e:3}
\revv{\boldsymbol{J}(k) \coloneqq \left[ \begin{array}{ccc} \frac{\partial f_1}{\partial p_1}(k) & \dots & \frac{\partial f_1}{\partial p_m}(k) \\ \vdots & \ddots & \vdots \\ \frac{\partial f_n}{\partial p_1}(k) & \dots & \frac{\partial f_n}{\partial p_m}(k) \end{array} \right],}
\end{equation}
and applying the chain rule in order to obtain the Jacobian \revv{of the decoupled structure, $\tilde{\boldsymbol{J}}(k)$ (again evaluated in $\boldsymbol{p}(k)$), we have that}
\begin{equation}
\label{e:Jac_d}
\revv{\tilde{\boldsymbol{J}}(k) = \boldsymbol{W} \operatorname{diag}\left(\begin{bmatrix} g'_1(z_1(k)) \quad \cdots \quad g'_r(z_r(k)) \end{bmatrix}\right) \boldsymbol{V}^{\top},}
\end{equation}
in which case $g'_i(z_i) \coloneqq \frac{\text{d}g_i(z_i)}{\text{d}z_i}$, \revv{and recalling that $\boldsymbol{z}(k) \coloneqq \boldsymbol{V}^{\top} \boldsymbol{p}(k)$}. Dreesen et al.\ found that the decoupling task can be solved by a simultaneous diagonalisation of a set of Jacobian matrices \cite{dreesen2014}. The idea relies on the tri-linear form which arises when collecting evaluations of \eqref{e:Jac_d} in a third dimension. The process is depicted on the right-hand side of Fig.~\ref{f:overview}. \revv{Stacking $N$ evaluations of $\tilde{\boldsymbol{J}}(k)$ in the third dimension expands the object into a three-way array $\tilde{\mathcal{J}}  \in \mathbb{R}^{n \times m \times N}$, i.e.\  $\tilde{\mathcal{J}}[:,:,k] \coloneqq \tilde{\boldsymbol{J}}(k)$}. Given the diagonal form \revv{\eqref{e:Jac_d}}, the collection of Jacobians may be written as a sum of $r$ outer products (or rank-one terms). Element-wise we have that,
\begin{equation}
\label{e:outer}
\revv{\tilde{\mathcal{J}}[o,\ell,k]= \sum_{i=1}^r \boldsymbol{W}[o,i]\boldsymbol{V}[\ell,i] \boldsymbol{G}'[k,i],}
\end{equation}
\revv{$\text{for}~o=1,\dots,n~\text{and}~\ell=1,\dots,m~\text{and}~k=1,\dots,N$. Here the matrix $\boldsymbol{G}' \in \mathbb{R}^{N \times r}$ stores the evaluations of the functions $g'_i(z_i)$, which appear along the central diagonal element of \eqref{e:Jac_d}, i.e.\ }
\begin{equation}
\label{e:def_G'}
\revv{\boldsymbol{G}'[k,i] \coloneqq g'_i(z_i(k)).}
 \end{equation}
A sum of rank-one terms defines a diagonal, \revv{or polyadic}, tensor decomposition \citep{kolda2009}. 
The bottom section of Fig.~\ref{f:overview} depicts the diagonal core tensor which is obtained when extracting the diagonal plane, \revv{$\boldsymbol{G}'$}\footnote{In the original work of Dreesen et al.\ \cite{dreesen2014}, and also in \cite{decuyper2021} $\boldsymbol{H}$ was used to denote $\boldsymbol{G}'$.}. It illustrates that the collection of Jacobians can be represented by three matrix factors, $\boldsymbol{W}$, $\boldsymbol{V}$, and $\boldsymbol{G}'$. 
The decomposed tensor may \revv{alternatively} be written in shorthand notation
\begin{equation}
\tilde{\mathcal{J}} \coloneqq \llbracket \boldsymbol{W}, \boldsymbol{V}, \boldsymbol{G}' \rrbracket.
\end{equation}


\begin{figure}
\begin{center}
\includegraphics[width=0.6\textwidth]{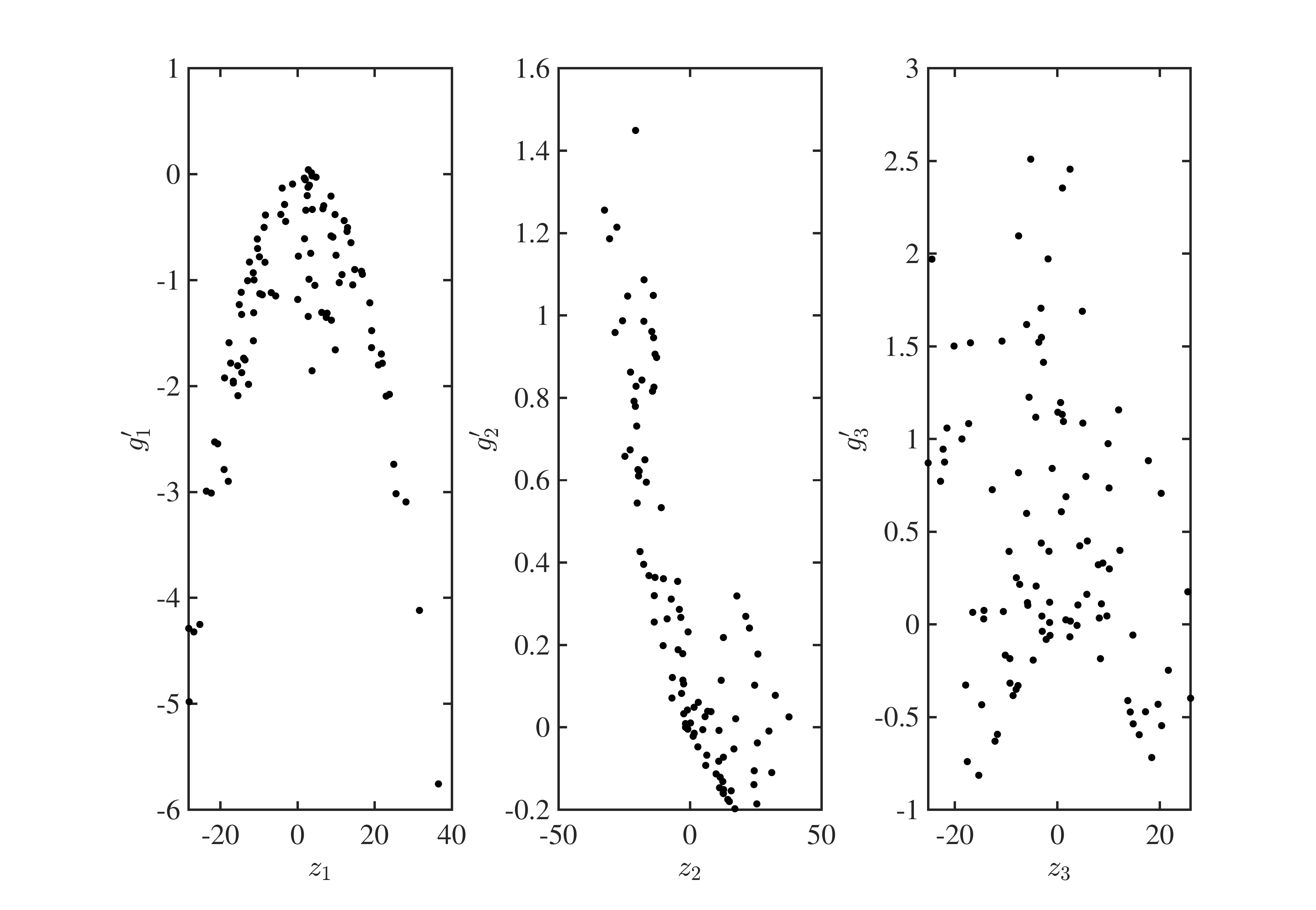}
\caption{\color{black} Visualisation of the nonparametric estimates of the functions $g'_i(z_i), i=1,\ldots,r$, which are obtained as the columns of matrix factor $\boldsymbol{G}' =\left[\boldsymbol{g}'_1, \ldots, \boldsymbol{g}'_r\right]$, see \ref{e:def_G'} for a definition of $\boldsymbol{G}'$. The result corresponds to the toy problem of Section \ref{s:toy} in which a third order polynomial, expressed in the standard monomial basis, is decoupled into three univariate branches. Although an exact solution exists, the decomposition is unable to convey information on the univariate functions, given that it is non-unique. Non-uniqueness of the tensor decomposition has been a limiting factor to the applicability of the method proposed by \cite{dreesen2014}. This limitation is overcome by the methods presented in this work.}
\label{f:2}
\end{center}
\end{figure}

The structure of $\tilde{\mathcal{J}}$ is the key to solving the decoupling problem. It implies that a coupled function may be mapped to a decoupled counterpart by factoring a three-way tensor storing first order derivative information. This leads to the following algorithm:
\begin{enumerate}
\item Collect the Jacobian matrices of the known coupled function, $\boldsymbol{J}$, and stack them into a three-way array, i.e.\ the Jacobian tensor $\mathcal{J}  \in \mathbb{R}^{n \times m \times N}$ (left-hand side of Fig.~\ref{f:overview}).
\item Compute a diagonal \revv{(polyadic)} tensor decomposition of $\mathcal{J}$, factoring it into $\{\boldsymbol{W}, \boldsymbol{V}, \boldsymbol{G}'\}$. This leads to the approximation ${\mathcal{J} \approx \tilde{\mathcal{J}}}$ and thus ${\mathcal{J} \approx  \llbracket \boldsymbol{W}, \boldsymbol{V},\boldsymbol{G}' \rrbracket}$, where the accuracy of the approximation relies on $r$. \revv{In the original method of \cite{dreesen2014}, the problem of exact decoupling was studied (referring to the equality in \eqref{e:2}). In that case also an exact tensor decomposition is required, hence the canonical polyadic decomposition (CPD) was used, which directly defines $r$ (see section \ref{s:notation}). In this work $r$ will be considered a design parameter, opening up the possibility of approximate decoupling.}
\end{enumerate}
\revv{Recalling the definition of $\boldsymbol{G}'$ in \eqref{e:def_G'}, the columns $\boldsymbol{g}'_1, \ldots, \boldsymbol{g}'_r$ store collections of evaluations of the function derivatives $g_i'(z_i)$. This means that $\boldsymbol{G}'$ provides nonparametric estimates of $g_i'(z_i)$.}
\begin{enumerate}
\setcounter{enumi}{2}
\item \revv{In a final step, the columns $\boldsymbol{g}'_i$ \revv{are parameterised as a function of $z_i$}. A parametric estimates of $g_i(z_i)$ may then be obtained through integration.}
\end{enumerate}
The method of \cite{dreesen2014} is compelling for the following reason: irrespective of the nature of the coupled function, or the size of the function in terms of $m$ and $n$, the procedure boils down to solving a third order tensor decomposition, \rev{whereas the complexity of alternative methods depend on the complexity of the original function, e.g.\ the degree of a multivariate polynomial \cite{usevich2020}.} It is, however, also subtle since it relies on two intrinsic assumptions:
\begin{enumerate}
\item The equality in \eqref{e:2}, which implies that the decoupling is exact. Whether this is attainable is dictated by the number of branches, $r$. In Section \ref{s:approx} we will show that decoupled functions of the form of \eqref{e:2} are universal approximators.
\item The uniqueness of the tensor decomposition into its factors, $\{ \boldsymbol{W}, \boldsymbol{V},\boldsymbol{G}'\}$, in order for the elements of $\boldsymbol{g}'_i$ to truly correspond to evaluations of $g'_i(z_i)$. Whether the decomposition is unique again relies on $r$ \cite{kruskal1989}.
\end{enumerate}
It turns out that the requirements on $r$ for unique decompositions and accurate decoupled structures can be conflicting, imposing strong limitations on the applicability of the method.  Fig.~\ref{f:2} illustrates a potential solution of $\boldsymbol{G}'$ in the case of a non-unique decomposition. The figure corresponds to the toy problem of Section \ref{s:toy} in which a third order polynomial, expressed in the standard monomial basis, is decoupled into three univariate branches. Although an exact solution exists, it is clear that the decomposition does not convey information on the univariate functions. In this work, a new method is presented, leveraging the powerful idea of decomposing the Jacobian tensor while tackling the challenges of approximate decoupling and non-unique decompositions. \rev{The proposed method enables to retrieve smooth estimates of decoupled functions in a nonparametric way.}

\revv{
\section{Notation}
\label{s:notation}
Vectors are denoted by lower-case bold-faced letters, e.g.\ $\boldsymbol{a} \in \mathbb{R}^p$. 
Matrices are denoted by bold-faced upper-case letters, e.g.\ $\boldsymbol{A}$. 
The columns of a matrix are denoted by bold-faced lower-case letters, e.g.\ we have $\boldsymbol{A} = \left[ \boldsymbol{a}_1 \quad \ldots \quad \boldsymbol{a}_r \right] \in \mathbb{R}^{p \times r}$. 
Tensors are denoted by calligraphic letters, e.g.\ $\mathcal{A} \in \mathbb{R}^{p \times r \times N}$. Matrix slices along the third dimension of a third order tensor are denoted by subscript indices, e.g.\ $\boldsymbol{A}_i$, alternatively square brackets are used, i.e.\ $\boldsymbol{A}_i \coloneqq \mathcal{A}{\left[:,:,i\right]}$.
Scalars are given lower case letters. The element of a vector $\boldsymbol{a}$, matrix $\boldsymbol{A}$ or a (third-order) tensor $\mathcal{A}$ may hence be denoted as $a_i$, $a_{ij}$ or $a_{ijk}$, respectively. Alternatively square brackets are used, e.g.\ $a_{ij} \coloneqq \boldsymbol{A}[i,j]$. A collection of $N$ unsorted elements, e.g.\ vectors $\boldsymbol{a}$, is denoted by $\left\{\boldsymbol{a}\left(k\right)\right\}_{k=1}^N$. The root mean square of a vector is denoted by $\text{rms}(\boldsymbol{a}) = \sqrt{\frac{1}{N} \sum_{i=1}^Na_i^2}$. The expected value is indicated by $\mathbb{E}$.

The Kronecker product of two matrices $\boldsymbol{A}$ and $\boldsymbol{B}$ is denoted $\boldsymbol{A} \kron \boldsymbol{B}$. 
The Khatri-Rao product of two matrices is denoted by $\boldsymbol{A} \kr \boldsymbol{B}$ \cite{kolda2009}. 
The Hadamard (element-wise) product of $\boldsymbol{A}$ and $\boldsymbol{B}$ is denoted by $\boldsymbol{A} * \boldsymbol{B}$.
The pseudo-inverse of a matrix $\boldsymbol{A}$ is denoted by $\boldsymbol{A}^\dagger$.
%
The vectorisation operator that converts a matrix or a tensor to a vector by concatenating its columns is denoted by $\operatorname{vec}({\boldsymbol{A}}) = \boldsymbol{a}$.
The mode-$n$ unfolding of a tensor $\mathcal{A}$ is denoted by $\boldsymbol{A}_{(n)}$. The columns of $\boldsymbol{A}_{(n)}$ are the mode-$n$ fibers (columns, rows, tubes) of $\mathcal{A}$ \cite{kolda2009}. CPD is used to denote the canonical polyadic decomposition of a tensor, i.e.\ a sum of the minimal number of rank-one terms for which the decomposition is exact. FTD is used to denote a filtered tensor decomposition, referring to the procedure which is developed in this work.}

\section{Problem statement}
\label{s:problem}
\rev{The objective is to have a procedure which allows one to approximate any arbitrary continuous multivariate function $\boldsymbol{f}(\boldsymbol{p})$ by a decoupled function. \revv{In practice $\boldsymbol{f}(\boldsymbol{p})$ is typically inferred from data during a training process. It is therefore justified to define the function approximation on the basis of $N$ operating points, drawn from the training distribution. This will ensure that the approximation is accurate in the domain of interest. The objective then assumes the form}
\begin{equation}
\label{e:4}
\underset{\boldsymbol{W},\boldsymbol{V},\boldsymbol{g} \in \mathcal{G}}{\operatorname{arg~min}} \frac{1}{N} \sum_{k=1}^N\lVert \boldsymbol{f}(\boldsymbol{p}(k))  - \boldsymbol{W} \boldsymbol{g}(\boldsymbol{V}^{\top} \boldsymbol{p}(k)) \rVert_2^2,
\end{equation}
with $\mathcal{G}$ a predefined function space. \revv{A direct approach to solving \eqref{e:4} would entail solving a hard nonlinear optimisation problem.} Instead, and similar to \cite{dreesen2014}, we will \revv{rely on the connection to tensor decompositions, observed when formulating the problem at the Jacobian level,}
\begin{equation}
\label{e:5}
\underset{\boldsymbol{W},\boldsymbol{V},\boldsymbol{G}'}{\operatorname{arg~min}}~\lVert \mathcal{J}  - \llbracket \boldsymbol{W}, \boldsymbol{V}, \boldsymbol{G}' \rrbracket \rVert_F^2.
\end{equation}
\revv{
Replacing \eqref{e:4} by \eqref{e:5} is not without consequence. Both the motivation and implications are listed below:
\begin{enumerate}
\item The main motivation is that \eqref{e:5} is tri-linear in the unknown factors. This opens up the problem to an alternating approach where by iteratively fixing two of the factors, the remaining problem of optimising the third factor admits a closed-form solution.
\item \eqref{e:5} is not constrained by a function family (this would destroy the multi-linearity). In other words, \eqref{e:def_G'} is not explicitly enforced. Instead, \cite{dreesen2014} relies on the uniqueness property of the tensor decomposition. In case of a unique decomposition, the factor $\boldsymbol{G}'$ will indeed store nonparametric estimates of the functions $g_i'(z_i)$, as given by \eqref{e:def_G'}.
\item Formulating the function approximation at the Jacobian level implies that the zeroth-order information is lost. An additional step will be required in order to recover the constant terms. This will be addressed in section \ref{s:DC}.
\item The solution obtained from \eqref{e:5} is not necessarily optimal with respect to \eqref{e:4}. In such cases a post-optimisation based on \eqref{e:5} can be used, where the solution obtained from the decoupling procedure serves as starting values.
\end{enumerate}}

\revv{The objective of this work is to have a procedure that no longer relies on the uniqueness property of the decomposition, which was found to be a limiting factor for the applicability of function decoupling. In order to be able to let go of a unique solution and still retrieve meaningful decoupled functions it is proposed to promote smoothness of the nonparametric function estimates. Promoting smoothness will prevent that non-meaningful solutions, similar to Fig.~\ref{f:2} are obtained. The benefit is twofold:
\begin{itemize}
\item By removing the requirement for uniqueness, $r$ can become a design parameter. Or in other words, the complexity of the decoupled target function can be tuned and balanced with respect to the accuracy of the approximation.
\item Promoting smoothness leads to meaningful results without having to constrain the function family. The proposed decoupling procedure therefore remains nonparametric, allowing any type of function to be handled.
\end{itemize}}


Smoothness is, however, not an objective that is to be quantified in itself. The level of required smoothness goes hand in hand with the final choice of basis expansion used to parametrise the branches, e.g.\ the degree of the polynomial basis. What will be quantified and monitored is the accuracy of the function approximation, for which a metric is introduced in \eqref{e:rel_er}, i.e.\ a normalised version of \eqref{e:4}.
}

\subsection{Contributions presented in this work}
Two new algorithms will be introduced. Both methods are distinct in how smoothness is achieved. In Section \ref{ss:implicit} smoothness is encoded implicitly in the objective. The result is a fully automated procedure requiring very little user interaction. In Section \ref{ss:explicit} regularisation is used to explicitly promote continuity of the derivative. An illustration will show that it is very intuitive to penalise non-smooth solutions. 

Alternative methods that aim at ensuring unique decompositions rely either on introducing information of the Hessian, or on introducing explicit polynomial constraints on the unknown $g_i$ \citep{karami2021,dreesen2018}. Note that such constraints do discriminate function families from the start. In this work a first-order approach is applied, avoiding the prohibitively expensive computation of the Hessian, while resorting to nonparametric methods to ensure smoothness, hence keeping the decomposition independent of any choice of basis expansion.

\section{Universal approximation theorem}
\label{s:approx}
It may be questioned under what conditions decoupled functions of the form of \eqref{e:2} can be used to approximate an arbitrary continuous multivariate function. In what follows we study the universal approximation properties of such decoupled forms.


\rev{\begin{thm}\label{thm1}
Let $A$ be the family of real continuous functions, defined on a compact set $K\subset\mathbb{R}^n$ that assume the form
\begin{equation}
   h_r(\boldsymbol{p}) = \sum_{i=1}^r w_i g_i(\boldsymbol{v}_i^{\top} \boldsymbol{p}),
\end{equation}
for some integer $r>0$, coefficients $w_i$, and vectors $\boldsymbol{v}_i\in\mathbb{R}^n$. Here, $g_i$ is a collection of univariate functions $g_i:\mathbb{R}\mapsto \mathbb{R}$. Let $C$ denote  the space of all real continuous functions on $K$. There exists a collection of univariate functions $g_i$ such that $A$ is dense in $C$ under the infinity or amplitude norm. That is, for any $f\in C$ and any $\epsilon >0$, there exists $r>0$, vectors $\boldsymbol{v}_i$ and coefficients $w_i$ such that $h_r\in A$ satisfies
\begin{equation}
 \| f-h_r \|_{\infty} \coloneqq \sup_{\boldsymbol{p}\in K} | f(\boldsymbol{p}) - h_r(\boldsymbol{p}) | < \epsilon.
 \end{equation}
\end{thm}
}

\revv{
\begin{pf} 
Consider the family of monomial interpolation functions $g_i(x)\coloneqq x^i$. The corresponding functions $h_r^{\star} \in A$ assume the form
\begin{equation}
h_r^{\star}(\boldsymbol{p}) = \sum_{i=1}^r w_i [\boldsymbol{v}_i^{\top}\boldsymbol{p} ] ^i. 
\end{equation}

Given that the monomials $[\boldsymbol{v}_i^{\top}\boldsymbol{p} ] ^i$ separate points of $K$ and are closed under multiplication, finite linear combinations of such functions form a point separating algebra which is dense in $C$ under the sup norm because of the Stone-Weierstrass theorem \cite{rudin1976}. \qed

\paragraph{Remark} Note that the choice of interpolation function is not restricted to monomials, e.g.\ the family of exponential functions $\sum_{i=1}^r w_i e^{\boldsymbol{v}_i^{\top} \boldsymbol{p}}$, with $w_i$ real, and $\boldsymbol{v}_i$ a vector of non-negative integers, would also do \cite{diaconis1984}. Moreover, observe that single-layer neural networks are a subset of the decoupled form. Therefore, the classically considered activation functions may also be considered \cite{cybenko1989}.
\end{pf}
}

\section{Nonparametric Filtered Tensor Decomposition (FTD)}
\label{s:FCPD}

The proposed idea to ensure smoothness revolves around introducing \emph{finite difference filters} into the decomposition routine of the Jacobian tensor. A finite difference filter is a matrix which upon multiplication with a vector of function evaluations, returns a finite difference approximation. The use of finite differences allows one to steer the decomposition towards meaningful results in a nonparametric way. 

Consider the couple of $N$ evaluations of the function $g_i(z_i)$, i.e.\ $\{z_{i}{(k)},g_{i}{(k)}\}_{k=1}^N$. Given that ${g'_i(z_i) \coloneqq \frac{\text{d}g_i(z_i)}{\text{d}z_i}}$, we can compute a finite difference approximation, ${\boldsymbol{g}'_i \in \mathbb{R}^N}$, by applying a finite difference filter to the vector of evaluations $\boldsymbol{g}_i$,
\begin{equation}
\label{e:hg}
\boldsymbol{g}'_i = \boldsymbol{S}_i^{-1} \boldsymbol{D}_i \boldsymbol{S}_i \boldsymbol{g}_i,
\end{equation}
where $\boldsymbol{S}_i$ is a sorting matrix, arranging the elements of $\boldsymbol{g}_i$ in ascending order of $z_i$, and $\boldsymbol{D}_i$ embodies the finite difference operator. Should $\boldsymbol{z}_i$ be equidistantly spaced, with a spacing of $\delta_{z_i}$, $\boldsymbol{D}_i$ could for instance be defined by the Toeplitz matrix
\begin{equation}
\label{e:D}
\boldsymbol{D}_i = \frac{1}{\delta_{z_i}} \left[ \begin{array}{ccccc} -1 & 1 & 0 &  \cdots & 0 \\ 0 & -1 & 1 & \ddots & \vdots \\ \vdots & \ddots & \ddots & \ddots & 0\\0 & \dots & 0 & -1 & 1 \end{array} \right],
\end{equation}
where this particular example is known as a 2-points forward differencing scheme (or a right-derivative approximation). In this context, the term \emph{finite difference filter}, is used to refer to the sequence of operations
\begin{equation}
\boldsymbol{F} \coloneqq \boldsymbol{S}^{-1} \boldsymbol{D} \boldsymbol{S}.
\end{equation}
In practice, $\boldsymbol{z}_i$, will not be equidistantly spaced. A generalisation of $\boldsymbol{D}$ is provided in \ref{a:1}. Introducing finite difference filters will prove beneficial in two ways:
\begin{itemize}
\item It will allow  $\mathcal{J}$ to be decomposed into $\{ \boldsymbol{W}, \boldsymbol{V},\boldsymbol{G}\}$, where $\boldsymbol{G}$ stores evaluations of the functions $g_i$, hence removing the need for an integration step (Section \ref{ss:G}).
\item It will provide the means to include a smoothness objective in the decomposition (Sections \ref{ss:implicit} and \ref{ss:explicit}) \citep{decuyper2019,decuyperFCPD2021}.
\end{itemize}

\subsection{A decomposition into $\boldsymbol{W}$,$\boldsymbol{V}$, and $\boldsymbol{G}$}
\label{ss:G}
Exploiting the relationship of \eqref{e:hg} we may propose an alternative factorisation of the Jacobian tensor where the third factor contains $\boldsymbol{G}$, i.e\ a matrix holding evaluations of the to be determined univariate functions $g_i(z_i)$ along its columns. This is in contrast to the factor $\boldsymbol{G}'$ which stores derivative information. The sought-after decomposition is then of the form
\begin{equation}
\label{e:G}
\mathcal{J} \approx \llbracket \boldsymbol{W}, \boldsymbol{V}, \mathcal{F}(\boldsymbol{V}) \circ \boldsymbol{G} \rrbracket,~\text{with}
\end{equation}
\begin{equation}
\label{e:H}
\boldsymbol{G}'\color{black} \coloneqq \color{black} \mathcal{F}(\boldsymbol{V}) \circ \boldsymbol{G}.
\end{equation}
The operation `$\circ$' symbolises the column-wise filtering of \eqref{e:hg}\footnote{This notation should not be confused with that of the outer product.}. For convenience, the finite difference filters, corresponding to each column, are stored in a three-way array $\mathcal{F} \in \mathbb{R}^{N \times N \times r}$ such that
\begin{equation} 
\mathcal{F}{[:,:,i]}(\boldsymbol{V}) \coloneqq \boldsymbol{F}_i(\boldsymbol{V}) = \boldsymbol{S}_i^{-1} \boldsymbol{D}_i \boldsymbol{S}_i.
\end{equation}
Notice the explicit dependence on the factor $\boldsymbol{V}$. The factor $\boldsymbol{V}$ defines the axes on the basis of which the finite difference is computed. Both the sorting matrix $\boldsymbol{S}_i$ and the finite difference operator $\boldsymbol{D}_i$ therefore depend on $\boldsymbol{V}$ (see \ref{a:1}). \eqref{e:H} may alternatively be written in vectorised form
\begin{equation}
\label{e:vec}
\operatorname{vec}\left(\boldsymbol{G}'\right) = \left[\begin{array}{cccc}\boldsymbol{F}_1 & 0 & \dots & 0 \\ 0 & \boldsymbol{F}_2 & \ddots & \vdots \\ \vdots & \ddots & \ddots & 0 \\ 0 & \cdots & 0 & \boldsymbol{F}_r \end{array} \right] \operatorname{vec}\left(\boldsymbol{G}\right).
\end{equation}
Given that the original decomposition $\{\boldsymbol{W},\boldsymbol{V},\boldsymbol{G}'\}$ has been translated to $\{\boldsymbol{W},\boldsymbol{V},\boldsymbol{G}\}$, also the smoothness requirement, which was formulated in Section \ref{s:problem}, is transferred onto $\boldsymbol{G}$. 

\subsection{Adding an implicit smoothness objective}
\label{ss:implicit}

Smoothness and finite differences are intimately linked, e.g.\ abrupt local changes will be emphasised by a finite difference filter. \revv{In \ref{a:1} a procedure to construct multiple finite difference filters which locally act on a different window of points, was introduced}. Given that \revv{all these} filters all approximate the first order derivative, differences amongst their results convey information on the local smoothness of the function.

\begin{figure*}
\begin{center}
\begin{subfigure}{0.32\textwidth}
\begin{center}
\includegraphics[width=0.96\textwidth]{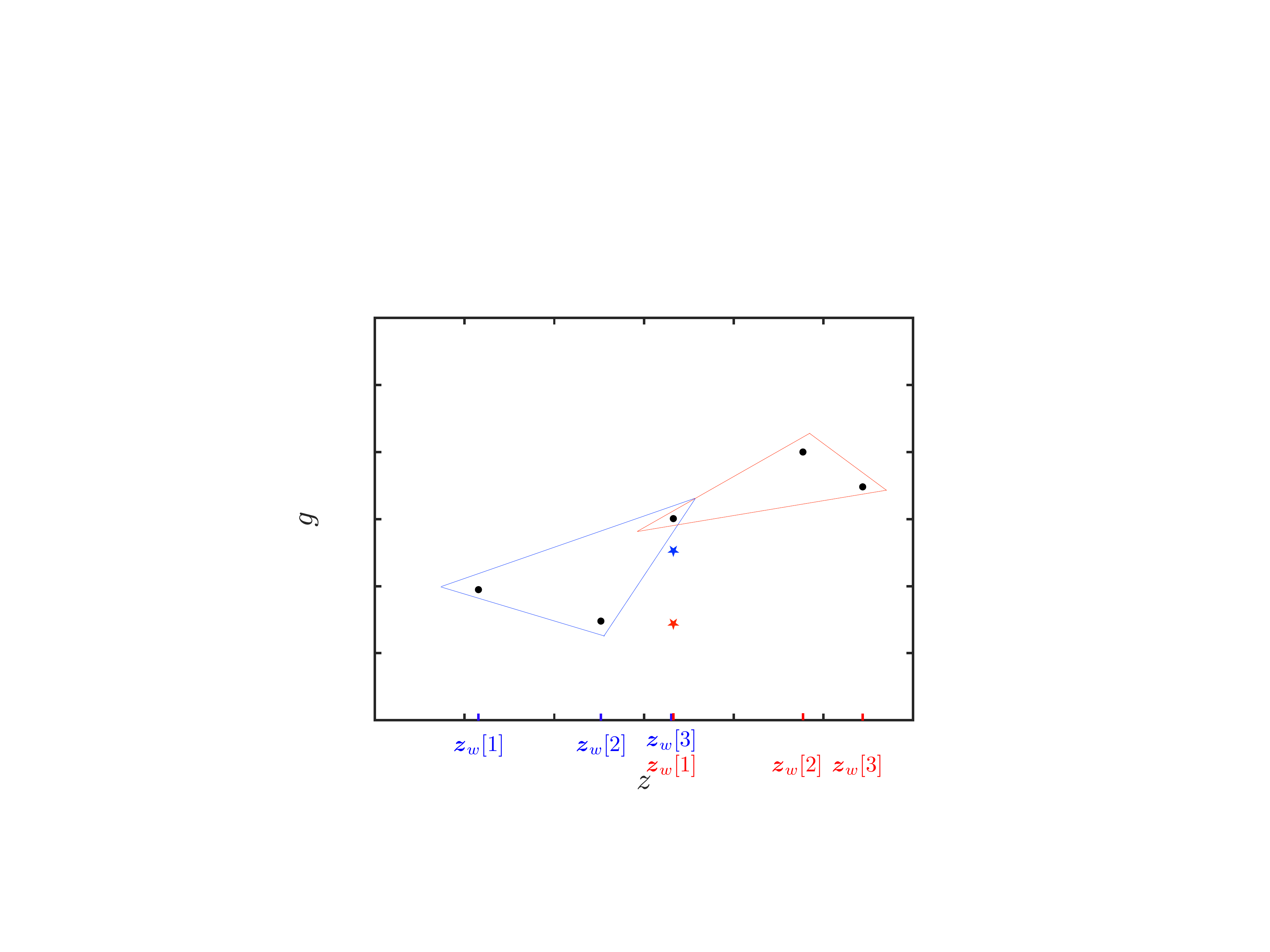}
\caption{}
\label{f:3a}
\end{center}
\end{subfigure}
\begin{subfigure}{0.32\textwidth}
\begin{center}
\includegraphics[width=\textwidth]{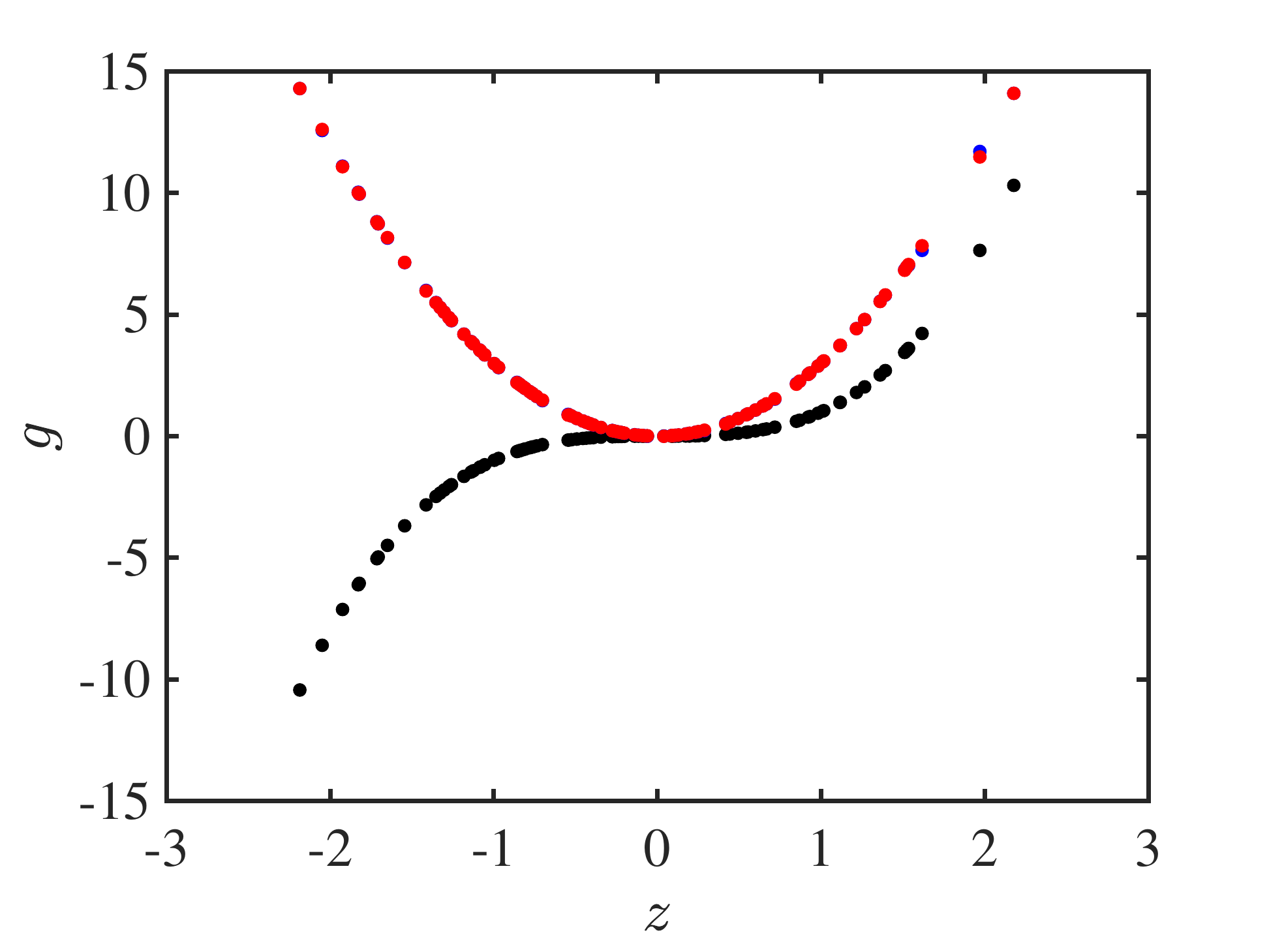}
\caption{}
\label{f:3b}
\end{center}
\end{subfigure}
\begin{subfigure}{0.32\textwidth}
\begin{center}
\includegraphics[width=\textwidth]{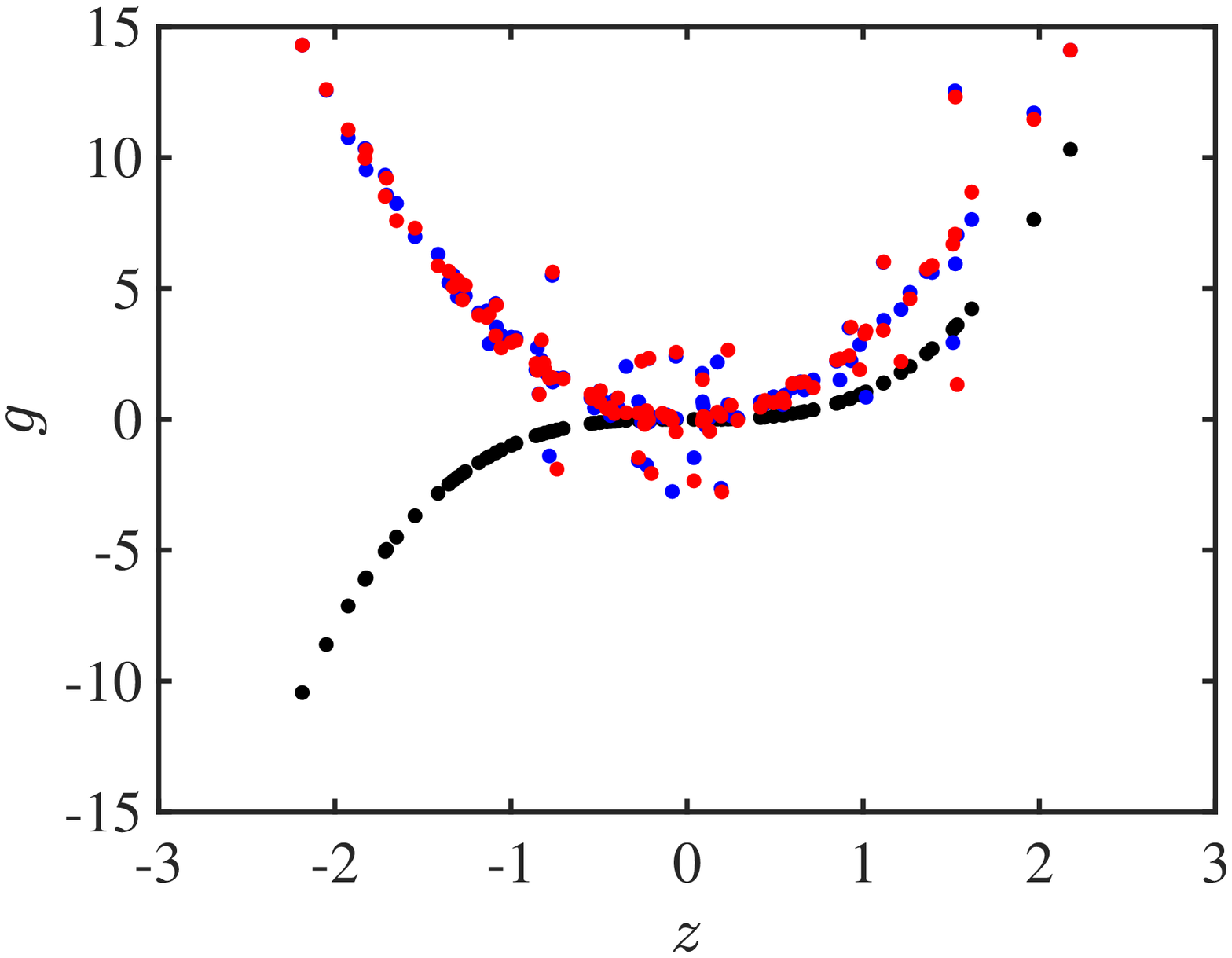}
\caption{}
\label{f:3c}
\end{center}
\end{subfigure}
\caption{(a) Black markers indicate a set of non-smooth evaluations $g(z)$. The blue and red $\star$ respectively denote the left and right 3-points finite difference, computed on the left and right window of points. (b) Black markers are evaluations of the smooth function $g = z^3$ for a set of randomly chosen points $\boldsymbol{z}$. Blue and red (almost on top of each other) are respectively the left and right 3-points finite differences. (c) Distorting the evaluations with Gaussian noise ($\sigma = 0.001\operatorname{rms}(\boldsymbol{g})$) results in clear differences amongst both finite difference approximations.}
\label{f:filter}
\end{center}
\end{figure*}

Fig.~\ref{f:3a} illustrates the operation of two distinct finite difference filters on a set of non-equidistantly spaced points. Both filters are created in accordance to \ref{a:1}. In blue, a \emph{3-points left finite difference} is used. The result is indicated with a blue $\star$. In red, a \emph{3-points right finite difference} is used (red $\star$). Both approximations yield a different result. This is due to local non-smoothness. Fig.~\ref{f:3b} and \ref{f:3c} demonstrate this property in an intuitive manner. Fig.~\ref{f:3b} depicts a number of evaluations of the smooth function $g=z^3$. In Fig.~\ref{f:3c}, these evaluations are distorted by Gaussian noise with $\sigma = 0.001 \operatorname{rms}(\boldsymbol{g})$. In both cases a 3-points left and a 3-points right finite difference is computed. The results are indicated by blue and red markers, respectively. While for the smooth function hardly any difference may be observed, the distorted evaluations clearly result in distinct derivative approximations. This implies that the deviation between left and right finite difference filters conveys information on the local smoothness of the evaluations. This property will be exploited in order to promote smooth solutions.

The example in Fig.~\ref{f:filter} alludes to the fact that decompositions of the form of \eqref{e:G}, which simultaneously satisfy multiple finite difference filters, must be smooth. In other words, considering multiple filters limits the number of possible solutions to a subset of smooth solutions. Smoothness can hence implicitly be encoded in the decomposition by constructing objective functions, for each matrix factor, on the basis of multiple filters.

Consider $s$ collections of finite difference filters, stored in the three-way arrays $\mathcal{F}_1$ to $\mathcal{F}_s$, each constructed on the basis of a different window of points (see \ref{a:1}). The claim is that only smooth solutions of $\boldsymbol{G}$ may satisfy 
\begin{equation}
\label{e:imp}
\mathcal{J} \approx \llbracket \boldsymbol{W}, \boldsymbol{V}, \mathcal{F}_1(\boldsymbol{V}) \circ \boldsymbol{G} \rrbracket \approx \ldots \approx  \llbracket \boldsymbol{W}, \boldsymbol{V}, \mathcal{F}_s(\boldsymbol{V}) \circ \boldsymbol{G} \rrbracket.
\end{equation}
In what follows we will introduce an algorithm able to solve the decomposition expressed by \eqref{e:imp}.

\subsubsection{Implicit algorithm}
A classical approach to solving a tensor decomposition is by tensor unfoldings in combination with an alternating least-squares (ALS) procedure. Denoting the unfolding of the third order tensor $\mathcal{J}$ along its rows, columns and tubes by $\boldsymbol{J}_{(1)}, \boldsymbol{J}_{(2)}, \boldsymbol{J}_{(3)}$, respectively, and using `$\kr$' to denote the Khatri-Rao product, we may write \citep{kolda2009}
\begin{subequations}
\begin{align}
\boldsymbol{J}_{(1)} &\approx \boldsymbol{W}((\mathcal{F}_i(\boldsymbol{V}) \circ \boldsymbol{G}) \kr \boldsymbol{V})^{\top}, \label{e:15a} \\ 
\boldsymbol{J}_{(2)} &\approx \boldsymbol{V}((\mathcal{F}_i(\boldsymbol{V}) \circ \boldsymbol{G}) \kr \boldsymbol{W})^{\top}, \label{e:15b}\\
\boldsymbol{J}_{(3)} &\approx (\mathcal{F}_i(\boldsymbol{V}) \circ \boldsymbol{G})(\boldsymbol{V} \kr \boldsymbol{W})^{\top}, \label{e:15c}
\end{align}
\end{subequations}
with $i=1,\ldots,f$.
In the first step of the procedure the factors $\{\boldsymbol{W}, \boldsymbol{V},\boldsymbol{G}\}$ are initialised with random numbers. Following the ALS approach \cite{carroll1970,harshman1970}, each matrix factor is then updated while treating both others as constants. This leads to the following objective functions and update formulas:

\begin{enumerate}
\item Update $\boldsymbol{W} \rightarrow \boldsymbol{W}^+$

From combining \eqref{e:imp} and \eqref{e:15a}, the following objective is inferred
\begin{equation}
\label{e:Wimp}
\underset{\boldsymbol{W}}{\operatorname{arg~min}}~\Vert \boldsymbol{J}_{(1)} - \boldsymbol{W}((\mathcal{F}_1(\boldsymbol{V}) \circ \boldsymbol{G}) \kr \boldsymbol{V})^{\top} \Vert_F^2
+ \ldots + \Vert \boldsymbol{J}_{(1)} - \boldsymbol{W}((\mathcal{F}_s(\boldsymbol{V}) \circ \boldsymbol{G}) \kr \boldsymbol{V})^{\top} \Vert_F^2.
\end{equation}
Since $\boldsymbol{W}$ appears linearly in the objective, an analytical update formula is obtained (\ref{a:B2}).
\item Update $\boldsymbol{V} \rightarrow \boldsymbol{V}^+$

Combining \eqref{e:imp} and \eqref{e:15b} leads to
\begin{equation}
\label{e:Vimp}
\underset{\boldsymbol{V}}{\operatorname{arg~min}}~\Vert \boldsymbol{J}_{(2)} - \boldsymbol{V}((\mathcal{F}_1(\boldsymbol{V}) \circ \boldsymbol{G}) \kr \boldsymbol{W})^{\top} \Vert_F^2
+ \ldots + \Vert \boldsymbol{J}_{(2)} - \boldsymbol{V}((\mathcal{F}_s(\boldsymbol{V}) \circ \boldsymbol{G}) \kr \boldsymbol{W})^{\top} \Vert_F^2.
\end{equation}
Given that $\boldsymbol{V}$ appears nonlinearly in the objective, nonlinear optimisation is required to compute an update $\boldsymbol{V}^+$. The optimisation is solved using a \revv{built-in MATLAB Levenberg-Marquardt solver} \cite{levenberg1944}. As initialisation a \revv{relaxation}, based on one of the filters is used. More specifically, the dependency of $\mathcal{F}_i$ on $\boldsymbol{V}$ is disregarded.  Denoting $\boldsymbol{G}'_i \coloneqq \mathcal{F}_i \circ \boldsymbol{G}$, we have that
\begin{equation}
\boldsymbol{V}_0^+ = \boldsymbol{J}_{(2)} \left(\left( \boldsymbol{G}'_i \kr \boldsymbol{W} \right)^{\top}\right)^{\dagger}.
\end{equation}
Using the property
 \begin{equation}
 (\boldsymbol{A} \kr \boldsymbol{B})^{\top}(\boldsymbol{A} \kr \boldsymbol{B}) = \left(\boldsymbol{A}^{\top}\boldsymbol{A}\right) * \left(\boldsymbol{B}^{\top}\boldsymbol{B}\right),
 \end{equation}
with `$*$' denoting the Hadamard product, a more efficient update formula may obtained, requiring only the inversion of an $r \times r$ matrix. The initialisation of the update then becomes
\begin{equation}
\boldsymbol{V}_0^+ = \boldsymbol{J}_{(2)} \left(\boldsymbol{G}'_i \kr \boldsymbol{W} \right)\left(\left({\boldsymbol{G}'_i}^{\top}\boldsymbol{G}'_i\right) * \left(\boldsymbol{W}^{\top} \boldsymbol{W}\right) \right)^{\dagger}.
\end{equation}
To improve conditioning every update is followed by a normalisation such that the columns are given by
$\frac{\boldsymbol{v}_i^+}{\Vert \boldsymbol{v}_i^+ \Vert_2}$.

\item Update $\boldsymbol{G} \rightarrow \boldsymbol{G}^+$

In an analogous way $\boldsymbol{G}$ is updated from combining  \eqref{e:imp} and \eqref{e:15c}, 
\begin{equation}
\label{e:Gimp}
\underset{\boldsymbol{G}}{\operatorname{arg~min}}~\Vert \boldsymbol{J}_{(3)} - (\mathcal{F}_1(\boldsymbol{V}) \circ \boldsymbol{G})(\boldsymbol{V} \kr \boldsymbol{W})^{\top} \Vert_F^2
+ \ldots + \Vert \boldsymbol{J}_{(3)} -  (\mathcal{F}_s(\boldsymbol{V}) \circ \boldsymbol{G})(\boldsymbol{V}  \kr \boldsymbol{W})^{\top} \Vert_F^2.
\end{equation}
It can be shown that $\boldsymbol{G}$ appears linearly in the objective, leading to an analytical update formula (\ref{a:B2}).
\end{enumerate}

Steps 1 to 3 are iterated until convergence or a stopping criterium is met. The updated factors are passed on from one iteration to the next. The univariate functions $g_i(z_i)$ are ultimately obtained from parameterising the columns $\boldsymbol{g}_i$ using a basis expansion which can be freely chosen by the user. 

The \emph{implicit} smoothness procedure results in a robust method for decoupling continuous multivariate functions, regardless of the function family. Moreover, only very little user interaction is needed. The general procedure allows one to include as many filters as required. In Section \ref{s:toy} we will show that, in practice, including two filters, e.g.\ a left and a right derivative filter, may be enough to ensure continuity of the derivative and hence smoothness of the solution. Additionally, the smoothness property may be relaxed by constructing filters which allow a finite number of discontinuities. Loosely said, the filters may be seen as moulds, shaping the functions into a desired form. One may for instance also envisage encoding saturating behaviour into the filters.   

\subsection{Adding an explicit smoothness objective}
\label{ss:explicit}
In the previous section, the solutions of the decomposition could be brought back to a smooth subset by requiring that the solution satisfies multiple filters. In an analogue way, multiple filters may also be used to explicitly express a smoothness objective.
 
Consider the following collections of filters: a collection of 3-points left finite difference filters, denoted by $\mathcal{F}_L$, 3-points right filters, denoted by $\mathcal{F}_R$ and 3-points central filters, denoted by $\mathcal{F}_C$. In analogy with \eqref{e:G}, the decomposition is of the form
\begin{equation}
\mathcal{J} \approx \llbracket \boldsymbol{W}, \boldsymbol{V}, \mathcal{F}_C(\boldsymbol{V}) \circ \boldsymbol{G} \rrbracket.  
\end{equation}
Using the standard unfoldings, the decomposition can be translated to the following equivalent expressions \cite{kolda2009}
\begin{subequations}
\begin{align}
\boldsymbol{J}_{(1)} &\approx \boldsymbol{W}((\mathcal{F}_C(\boldsymbol{V}) \circ \boldsymbol{G}) \kr \boldsymbol{V})^{\top}, \label{e:15d} \\ 
\boldsymbol{J}_{(2)} &\approx \boldsymbol{V}((\mathcal{F}_C(\boldsymbol{V}) \circ \boldsymbol{G}) \kr \boldsymbol{W})^{\top}, \label{e:15e}\\
\boldsymbol{J}_{(3)} &\approx (\mathcal{F}_C(\boldsymbol{V}) \circ \boldsymbol{G})(\boldsymbol{V} \kr \boldsymbol{W})^{\top}. \label{e:15f}
\end{align}
\end{subequations}
When inferring objective functions from \eqref{e:15d} up to \eqref{e:15f}, appropriate penalty terms will be added using regularisation. \revv{In this work we will make use of regularisation to penalise divergent results from a left and a right finite difference filter. Minimising the difference between the left and the right derivative favours differentiable functions (and therefore also continuity). This will steer the decomposition towards smooth solutions. It must be stressed that other penalty terms, reflecting possibly other desired properties, may also be included.}

\subsubsection{Explicit algorithm}
The factors $\{\boldsymbol{W}, \boldsymbol{V},\boldsymbol{G}\}$ are initialised with random numbers. Also here an ALS approach is adopted. This leads to the following objective functions and update formulas:
\begin{enumerate}
\item Update $\boldsymbol{W} \rightarrow \boldsymbol{W}^+$

From Eq.\eqref{e:15d} we infer the following objective
\begin{equation}
\underset{\boldsymbol{W}}{\operatorname{arg~min}}~\Vert \boldsymbol{J}_{(1)} - \boldsymbol{W}((\mathcal{F}_C(\boldsymbol{V}) \circ \boldsymbol{G}) \kr \boldsymbol{V})^{\top} \Vert_F^2.
\end{equation}
For convenience we will denote ${\boldsymbol{G}'_C \coloneqq \mathcal{F}_C(\boldsymbol{V}) \circ \boldsymbol{G}}$. Given that $\boldsymbol{W}$ appears linearly in the objective, an analytical update formula is obtained,
\begin{equation}
\boldsymbol{W}^+ = \boldsymbol{J}_{(1)} \left(\left( \boldsymbol{G}'_C \kr \boldsymbol{V} \right)^{\top}\right)^{\dagger}.
\end{equation}
Using the property
 \begin{equation}
 (\boldsymbol{A} \kr \boldsymbol{B})^{\top}(\boldsymbol{A} \kr \boldsymbol{B}) = \left(\boldsymbol{A}^{\top}\boldsymbol{A}\right) * \left(\boldsymbol{B}^{\top}\boldsymbol{B}\right),
 \end{equation}
leads to the more efficient update formula, requiring only the inversion of an $r \times r$ matrix
\begin{equation}
\boldsymbol{W}^+ = \boldsymbol{J}_{(1)} \left(\boldsymbol{G}'_C \kr \boldsymbol{V} \right)\left(\left({\boldsymbol{G}'_C}^{\top}\boldsymbol{G}'_C\right) * \left(\boldsymbol{V}^{\top} \boldsymbol{V}\right) \right)^{\dagger}.
\end{equation}
Notice that the smoothness of $\boldsymbol{G}$ does not depend on $\boldsymbol{W}$. 
\item Update $\boldsymbol{V} \rightarrow \boldsymbol{V}^+$

Apart from the tensor approximation objective, which is implied by \eqref{e:15e}, the update of $\boldsymbol{V}$ must also take into account the smoothness of $\boldsymbol{G}$. The smoothness is indeed affected by $\boldsymbol{V}$ since $\boldsymbol{V}$ defines the $z$-axis, $\boldsymbol{z} = \boldsymbol{V}^{\top}\boldsymbol{p}$ (see Fig.~\ref{f:2}). The smoothness objective is therefore explicitly added using regularisation,
\begin{equation}
\label{e:V}
\underset{\boldsymbol{V}}{\operatorname{arg~min}}~\Vert \boldsymbol{J}_{(2)} - \boldsymbol{V}((\mathcal{F}_C(\boldsymbol{V}) \circ \boldsymbol{G}) \kr \boldsymbol{W})^{\top} \Vert_F^2 + \lambda \Vert \left(\mathcal{F}_L(\boldsymbol{V}) \circ \boldsymbol{G} \right) - \left(\mathcal{F}_R(\boldsymbol{V}) \circ \boldsymbol{G} \right) \Vert_F^2,
\end{equation}
where $\lambda$ is a hyperparameter which balances both objectives. The additional term penalises divergent results from a left and a right finite difference filtering operation, ultimately steering the optimisation towards smooth solutions. \revv{The penalty term in \eqref{e:V} was preferred over classical Tikhonov regularisation \cite{navasca2008}, i.e. a penalty of the form $\| \mathcal{F}_C \circ \boldsymbol{G}\|_F^2$, which was found to have the tendency to overly flatten the function estimates.}

Given that $\boldsymbol{V}$ appears nonlinearly in \eqref{e:V}, nonlinear optimisation is required to obtain $\boldsymbol{V}^+$. In order to promote smoothness evenly on all the columns of $\boldsymbol{G}$, a normalisation is needed. Denoting ${\boldsymbol{G}'_L \coloneqq \mathcal{F}_L \circ \boldsymbol{G}}$, and ${\boldsymbol{G}'_R \coloneqq \mathcal{F}_R \circ \boldsymbol{G}}$, all columns are normalised by their rms value, i.e.\ $\boldsymbol{g}'_{L_i}/\text{rms}(\boldsymbol{g}'_{L_i})$, and $\boldsymbol{g}'_{R_i}/\text{rms}(\boldsymbol{g}'_{R_i})$.

The optimisation is solved using a \revv{built-in MATLAB Levenberg-Marquardt solver \cite{levenberg1944}. For an analytical approach to the optimisation problem, the reader is referred to \cite{renczes2022}}. As initialisation \revv{a relaxation is used (i.e.\ without penalty term and disregarding the dependency of $\mathcal{F}_C$ on $\boldsymbol{V}$)}
\begin{equation}
\boldsymbol{V}_0^+ = \boldsymbol{J}_{(2)} \left(\boldsymbol{G}'_C \kr \boldsymbol{W} \right)\left(\left({\boldsymbol{G}'_C}^{\top}\boldsymbol{G}'_C\right) * \left(\boldsymbol{W}^{\top} \boldsymbol{W}\right) \right)^{\dagger}.
\end{equation}
To improve conditioning every update is followed by a normalisation such that the columns are given by
$\frac{\boldsymbol{v}_i^+}{\Vert \boldsymbol{v}_i^+ \Vert_2}$.

\item Update $\boldsymbol{G} \rightarrow \boldsymbol{G}^+$

In analogy with \eqref{e:V}, the update formula of $\boldsymbol{G}$ is also found from a joint objective function.
\begin{equation}
\label{e:G_update}
\underset{\boldsymbol{G}}{\operatorname{arg~min}}~\Vert \boldsymbol{J}_{(3)} -(\mathcal{F}_C(\boldsymbol{V}) \circ \boldsymbol{G})(\boldsymbol{V} \kr \boldsymbol{W})^{\top} \Vert_F^2 + \lambda \Vert \left(\mathcal{F}_L(\boldsymbol{V}) \circ \boldsymbol{G} \right) - \left(\mathcal{F}_R(\boldsymbol{V}) \circ \boldsymbol{G} \right) \Vert_F^2.
\end{equation}
It can be shown that $\boldsymbol{G}$ appears linearly in \eqref{e:G_update}. This allows for an analytical update formula (given in \ref{a:B1}).
\end{enumerate}

Steps 1 to 3 are iterated until convergence is reached or a stopping criterium is met. The updated factors are passed on from one iteration to the next. The appropriate value of $\lambda$ is to be determined from a line search. The univariate functions $g_i(z_i)$ are ultimately obtained from parameterising the columns $\boldsymbol{g}_i$ using a basis expansion which can be freely chosen by the user. 

Explicitly penalising divergent results from left and right finite difference filters is an intuitive way of introducing smoothness into the objective. An illustration in Section \ref{s:toy} will demonstrate that the level of smoothness responds as anticipated to the choice of the hyperparameter $\lambda$. It allows for a controlled balancing between accuracy of the decomposition and smoothness. The added flexibility comes at the price of increased computing times when scanning over $\lambda$.

%

\section{Estimating the constant terms}
\label{s:DC}

Inherent to decoupling on the basis of first-order derivative information is the loss of the constant terms. The constant term of every function $f_i$ of $\boldsymbol{f}\left(\boldsymbol{p}\right): \mathbb{R}^m \mapsto \mathbb{R}^n $ may, however, be directly estimated from input-output data. Let $\Delta_f \in \mathbb{R}^{N \times n}$ denote the error between $\boldsymbol{f}$ and the retrieved decoupled function $\boldsymbol{f}_D$, computed on the operating points such that
\begin{equation}
\Delta_f[i,j] = f_j\left(\boldsymbol{p}(i)\right) - f_{D_{j}}\left(\boldsymbol{p}(i)\right),
\end{equation}
with $j=1,\ldots,n$ and $i=1,\ldots,N$. A least-squares estimate of the constant terms, $\boldsymbol{c} \in \mathbb{R}^n$, is then found from
\begin{equation}
\boldsymbol{c}^{\top}= \left(\boldsymbol{1}_N^{\top} \boldsymbol{1}_N \right)^{-1} \boldsymbol{1}_N^{\top} \Delta_f,
\end{equation}
with $\boldsymbol{1}_N \in \mathbb{R}^N$ a vector of ones.

The constant terms may be incorporated in $\boldsymbol{f}_D =\boldsymbol{W} \boldsymbol{g}\left(\boldsymbol{V}^{\top} \boldsymbol{p} \right)$ by adding an additional branch
\begin{equation}
\tilde{\boldsymbol{f}}_D = \left[\boldsymbol{W} \quad \boldsymbol{c} \right] \begin{bmatrix} g_1(z_1) \\ \vdots \\ g_r(z_r) \\ 1 \end{bmatrix}.
\end{equation}

\section{Case study 1: a polynomial example}
\label{s:case+poly}

The objective is to demonstrate the capability of the method to return decoupled functions irrespective of the tensor decomposition properties (i.e.\ for non-exact and non-unique decompositions). The first example is of the polynomial type. In Section \ref{s:case_neural} a neural network will be studied. Recall that, given the universal approximation theorem of Section \ref{s:approx}, any continuous function may be approximated by a decoupled form.

\subsection{Toy problem}
\label{s:toy}
Both the explicit procedure of Section \ref{ss:explicit} and the implicit method of Section \ref{ss:implicit} will first be demonstrated on a toy problem.
Consider the decoupled polynomial function
\begin{subequations}\label{e:dec}
\begin{align}
\boldsymbol{f}_D(\boldsymbol{p}) &= \overbrace{\left[ \begin{array}{ccc}  3 & 0.5 & -1 \\ 1& 2 & 3 \end{array} \right]}^{\boldsymbol{W}} \overbrace{\left[ \begin{array}{c}  z_1^3 + 0.5 z_1^2 \\ 2 z_2^3 + z_2^2 \\ z_3^3 + 3 z_3^2 \end{array} \right]}^{\boldsymbol{g}(\boldsymbol{z})} \\
\boldsymbol{z}&= \overbrace{\left[ \begin{array}{cc}  1 & 2 \\ 3 & 1 \\ 0.5 & 3 \end{array}\right]}^{\boldsymbol{V}^{\top}} \boldsymbol{p},
\end{align}
\end{subequations} 
for which $m=2$, $n=2$ and $r=3$. An equivalent\footnote{The equivalency is subject to rounding errors in the computation of the coefficients.} formulation, expressed in the standard monomial basis reads
\begin{equation}\label{e:coupled}
\boldsymbol{f}(\boldsymbol{p}) = \left[ \begin{array}{ccccccc}  5.25 & 0 & -20.5 & 29.875 & 42.75 & 31.5 & -2 \\ 20.75 & 41 & 85 & 109.375 & 120.75 & 88.5 & 93 \end{array} \right] \begin{bmatrix} p_1^2 \\ p_1p_2 \\ p_2^2 \\ p_1^3 \\ p_1^2 p_2 \\ p_1 p_2^2 \\ p_2^3 \end{bmatrix}.
\end{equation}
One may think of \eqref{e:coupled} as a standalone static multivariate polynomial although it might as well be the internal part of a dynamical model. It could for instance be describing a polynomial NARX model \cite{billings2013}, in which case $p_1$ and $p_2$ would represent (delayed) input or output samples and $n$ would be equal to 1. Furthermore, it could be the nonlinear function which describes the state update in nonlinear state-space models, in which case $p_1$ and $p_2$ would be either state or input variables \citep{decuyper2021,csurcsia2022}.

\begin{figure}
\begin{center}
\begin{subfigure}[b]{0.45\textwidth}
\begin{center}
\includegraphics[width=\textwidth]{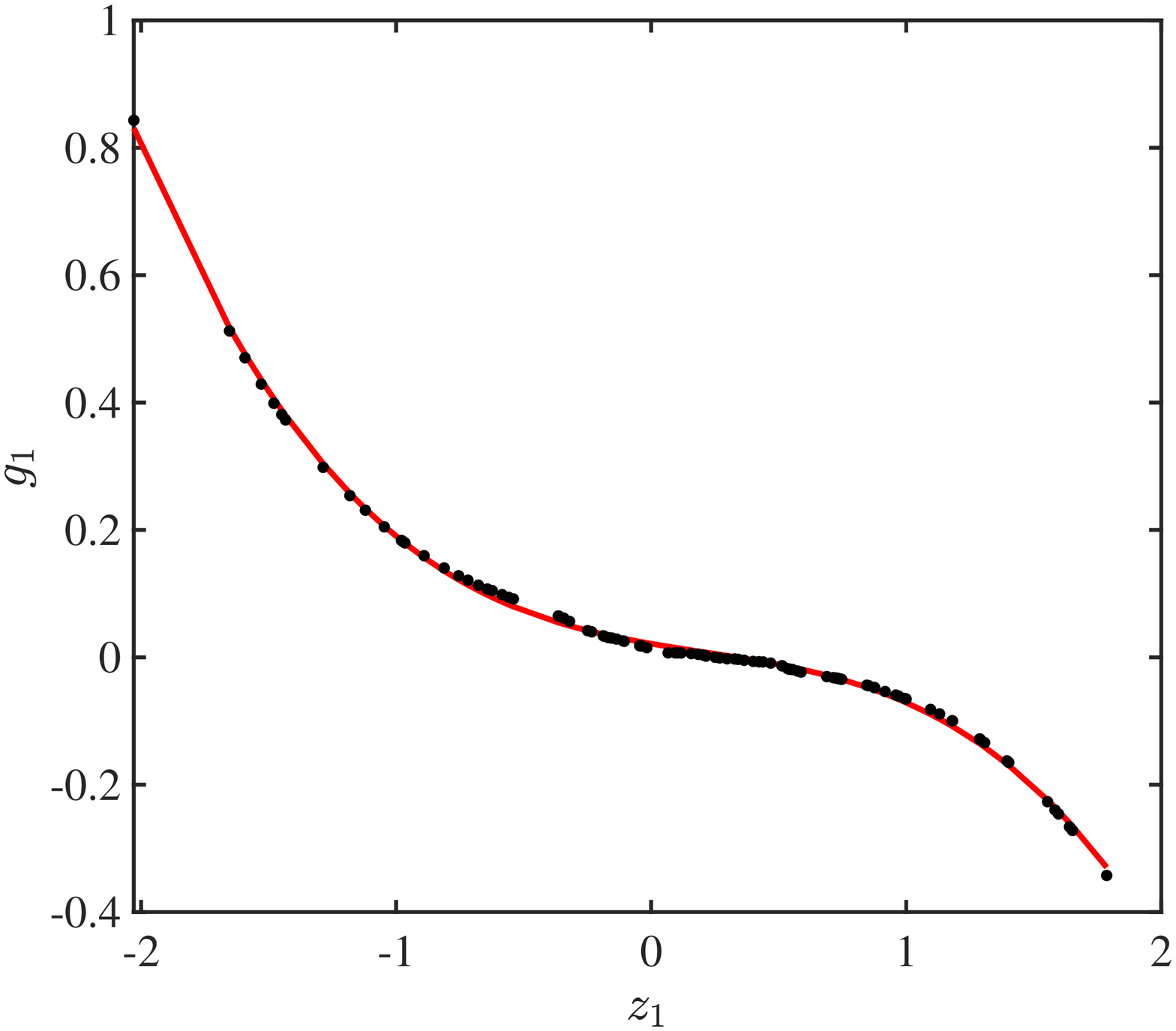}
\caption{}
\label{f:5a}
\end{center}
\end{subfigure}
\begin{subfigure}[b]{0.45\textwidth}
\begin{center}
\includegraphics[width=\textwidth]{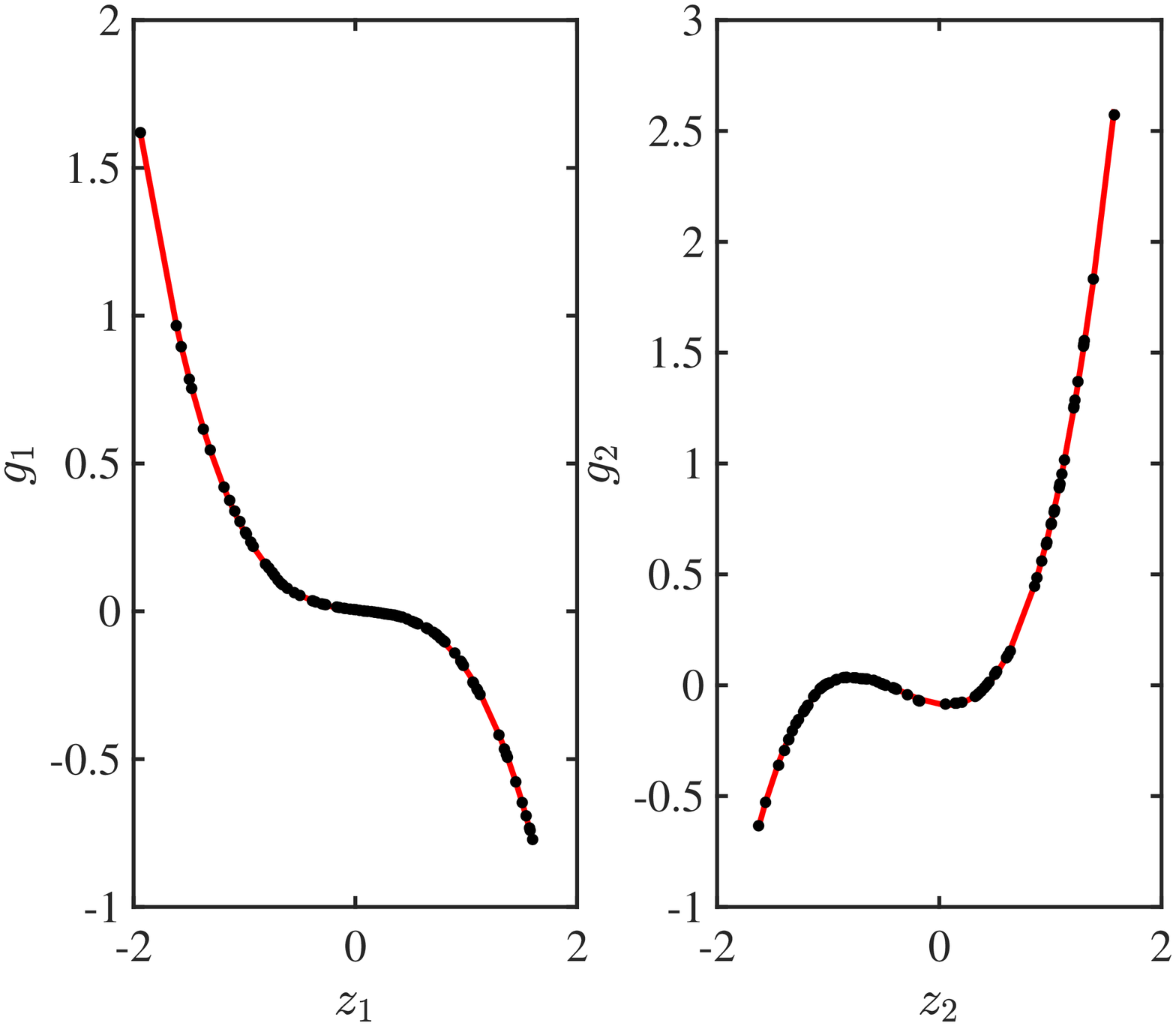}
\caption{}
\label{f:5b}
\end{center}
\end{subfigure}
\begin{subfigure}[b]{0.8\textwidth}
\begin{center}
\includegraphics[width=0.9\textwidth]{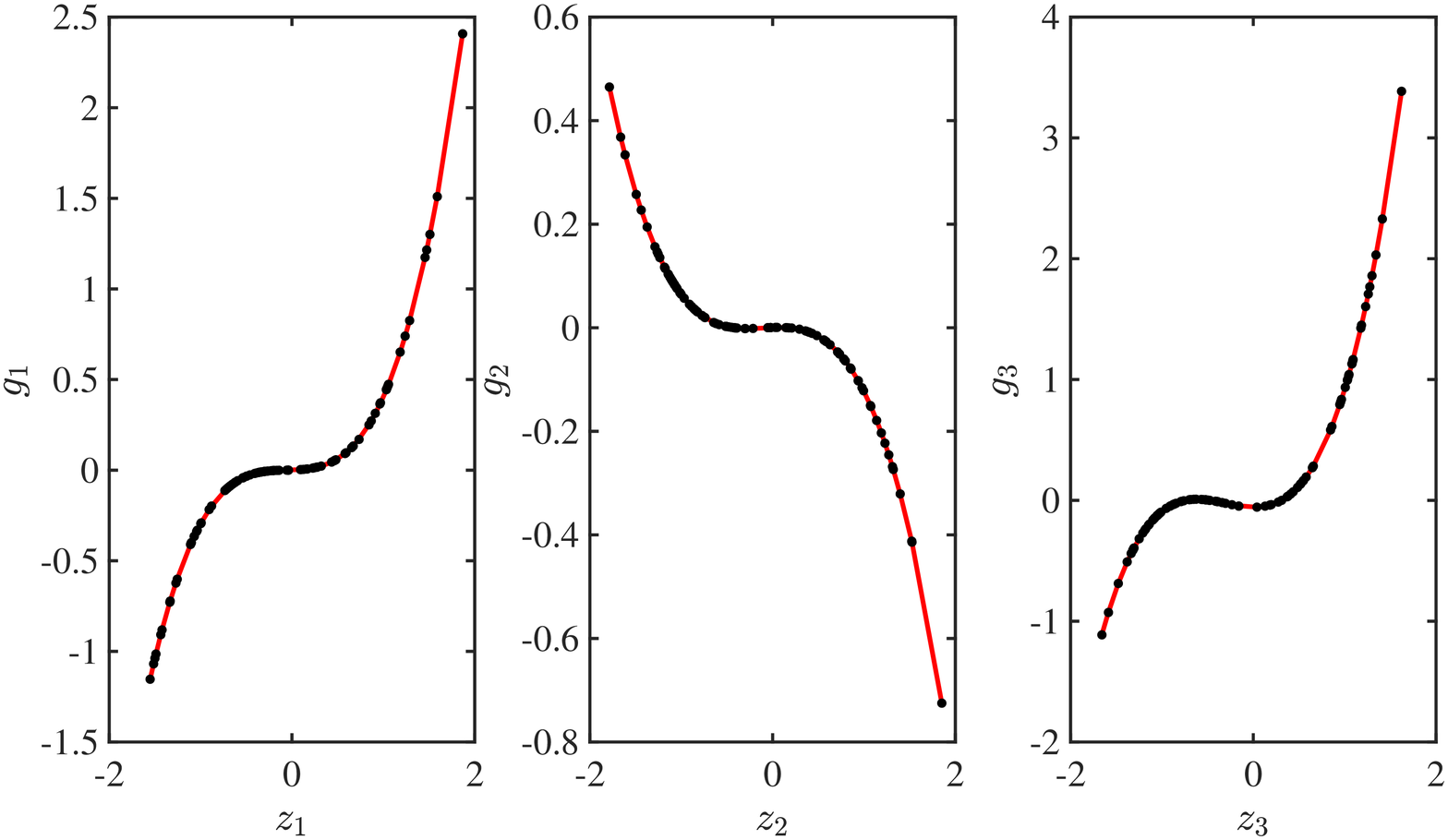}
\caption{}
\label{f:5c}
\end{center}
\end{subfigure}
\begin{subfigure}[b]{\textwidth}
\begin{center}
\includegraphics[width=\textwidth]{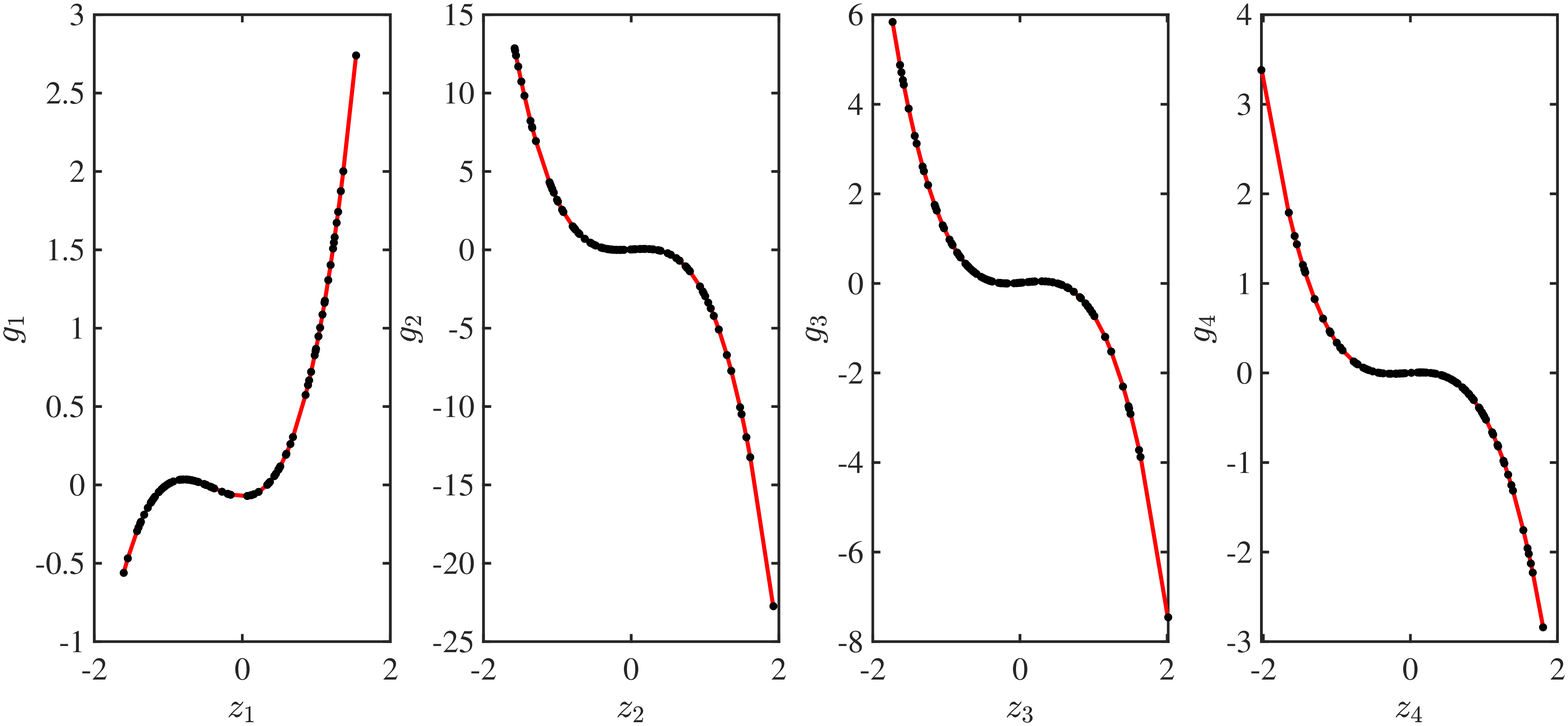}
\caption{}
\label{f:5d}
\end{center}
\end{subfigure}
\vspace{-0.75cm}
\caption{Results on decoupling the toy problem using the implicit FTD method, based on the use of two filters: a 3-points left finite difference and a 3-points right finite difference. (a) $r=1$ (b) $r=2$ (c) $r=3$ (d) $r=4$. The nonparametric estimates are fitted using polynomials of the third degree. Relative function errors can be consulted in Table \ref{t:1}.}
\label{f:Imp_r}
\end{center}
\end{figure}

\begin{figure}
\begin{subfigure}[b]{\textwidth}
\begin{center}
\includegraphics[width=0.7\textwidth]{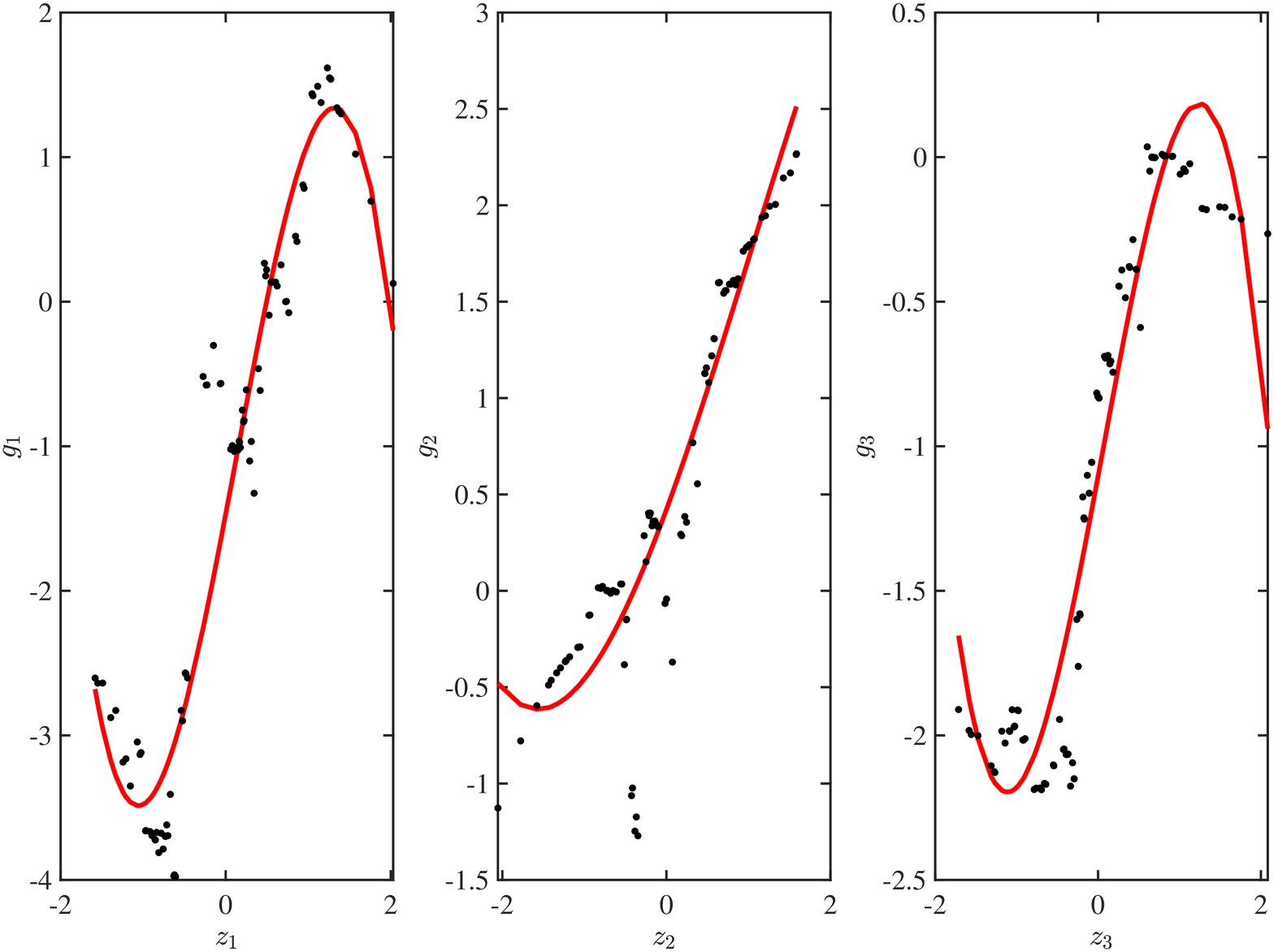}
\caption{}
\label{f:4a}
\end{center}
\end{subfigure}
\begin{subfigure}[b]{\textwidth}
\begin{center}
\includegraphics[width=0.7\textwidth]{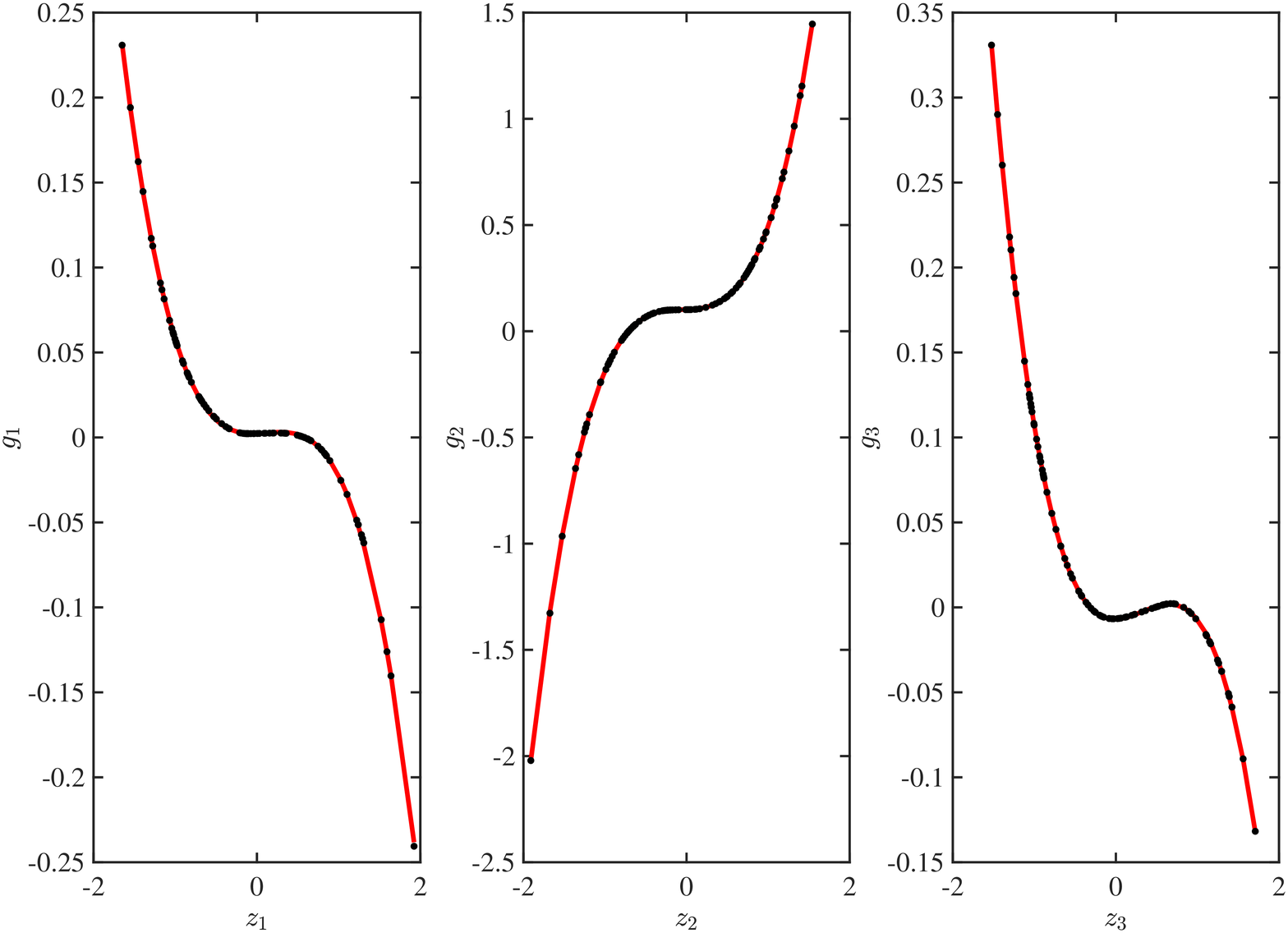}
\caption{}
\label{f:4b}
\end{center}
\end{subfigure}
\caption{Results on decoupling the toy problem using the explicit FTD method. (a) Selecting $\lambda = 0.01$ does not put enough weight on the smoothness objective. (b) A value of $\lambda = 10^{10}$ results in smooth estimates (black markers). In practice $\lambda$ is found from a line search. Red corresponds to a polynomial fit of the third degree.}
\label{f:lambda}
\end{figure}

The objective is to retrieve the decoupled function of the form \eqref{e:dec} starting from \eqref{e:coupled}. 
\begin{enumerate}
\item The first step consists in evaluating the Jacobian of \eqref{e:coupled}, $\boldsymbol{J}$, and stacking the matrices in the third dimension. A number of $N=100$ operating points were randomly selected from $\mathcal{U}\left(-1.5,1.5\right)$, leading to $\mathcal{J} \in \mathbb{R}^{2 \times 2 \times 100}$. In practice the operating points should cover the domain of interest.
\item The next step involves decomposing the Jacobian tensor. At this stage the number of univariate branches that make up the decoupled function, are selected. By construction we know that for the given example a solution containing $r=3$ branches exists (\eqref{e:dec}). 
\begin{enumerate}
\item \textbf{CPD}:
First we examine the classical canonical polyadic decomposition (CPD), which returns
 ${\mathcal{J} \approx \llbracket \boldsymbol{W}, \boldsymbol{V}, \boldsymbol{G}' \rrbracket}$. With a tensor approximation (\eqref{e:5}) close to machine precision, the decomposition was successful. Studying the result, however, reveals an underlying problem (depicted in Fig.~\ref{f:2}). As may be observed, the decomposition did not convey information on the univariate functions, which make up \eqref{e:dec}. The decomposition only points to \eqref{e:dec} in case it is unique, which is not the case in this example for $r=3$.

\item  \textbf{FTD implicit}: Not all solutions of the decomposition prove useful in our search for the decoupled function. We can, however, limit the possible solutions to a smooth subset by using an objective based on multiple finite difference filters, i.e.\ using the implicit algorithm of Section \ref{ss:implicit}. Given that in this case, $r$ is a design parameter, we can study the solution for a range of $r$ irrespective of whether an exact decoupled form exists. We will construct the objective (\eqref{e:imp}) on the basis of two filters, a 3-points left finite difference and a 3-points right finite difference. The procedure is then fully automated. The results for a scan of $r=1$ to 4 are presented in Fig.~\ref{f:Imp_r}.

\item  \textbf{FTD explicit}: Using the explicit algorithm of Section \ref{ss:explicit} we can steer the decomposition towards appropriate solutions by penalising non-smooth behaviour. Also in this case $r$ is a design parameter. Fig.~\ref{f:lambda} depicts the role of the hyperparameter $\lambda$ for decompositions with $r=3$.

\end{enumerate}
 \item To finalise the decoupling procedure, the estimates of the functions $g_i$ are parameterised. At this stage any basis expansion which deems fit can be used, provided that it is linear in the parameters. In Fig.~\ref{f:Imp_r} and \ref{f:4b} accurate third order polynomials were fitted (red curves). To assess the accuracy of the decoupled function, a relative root-mean-squared error is computed on the operating points
 \begin{equation}
 \label{e:rel_er}
 e_i = \frac{\sqrt{\frac{1}{N}\sum_{k=1}^N\left(f_i(\boldsymbol{p}(k)-f_{D_i}(\boldsymbol{p}(k))\right)^2}}{\sqrt{\frac{1}{N}\sum_{k=1}^N \left(f_i(\boldsymbol{p}(k))-\mathbb{E}\left(f_i\right)\right)^2}} \times 100,
 \end{equation}
with $\mathbb{E}\left(f_i\right)$ the expected value over all $N$ points. The results are listed in Table \ref{t:1}. In case of the explicit method, the lowest error found from a coarse search over $\sqrt{\lambda} = 10^{-1},10^0,10^1,10^2,10^3,10^4$, is reported. 
 
 Accurate decoupled approximations of \eqref{e:coupled} were retrieved. Note that $r$ can be used to balance model complexity to accuracy. The exact decoupled function, \eqref{e:dec} is, however, not retrieved given that the ALS algorithm leads to a local minimum of the objectives. Adding additional flexibility, i.e.\ selecting $r=4$ (while an exact $r=3$ solution exists) results in errors below 1\%. Both methods yield accurate decoupled functions. The implicit method has the advantage of being fully automated and requiring a significantly lower computing time. The computing time of the explicit method scales with the considered range of $\lambda$\footnote{The scan over $\lambda$ can be parallelised, reducing computing time.}. The accuracy could potentially be increased by using a finer search grid of $\lambda$.
 \begin{table}[h]
\caption{Relative root-mean-squared error on the function approximation of the toy problem.}
\label{t:1}
\begin{center}
\renewcommand*{\arraystretch}{1.2}
\begin{tabular}{| c  c | c | c | c | c |}
\cline{3-6}
\multicolumn{2}{c|}{} & $r=1$ & $r=2$ & $r=3$ & $r=4$   \\
\hline
\multirow{2}{*}{Implicit} & $e_{{1}}$ & 51.4 \% & 20.9 \% & 0.8 \% & 0.3 \%   \\ 
& $e_{{2}}$ & 32.1 \% & 6.2 \% & 1.0 \% & 0.4 \%    \\
\hline
\hline
\multirow{2}{*}{Explicit} & $e_{{1}}$ & 60.7 \% & 22.8 \% & 2.8 \% & 0.2 \%   \\
& $e_{{2}}$ & 32.6 \% & 4.9 \% & 1.3 \% & 0.1 \%    \\
\hline
\end{tabular}
\end{center}
\end{table}
\end{enumerate}

\begin{con}
Both the implicit and the explicit method return accurate decoupled functions. The implicit method has the advantage of being fully automated. It was shown that the use of two filters, a 3-points left finite difference and a 3-points right finite difference, is sufficient for smooth solutions to be obtained.
\end{con}


\subsection{Polynomial reduction}
\label{ss:poly_reduction}

\begin{figure}
\begin{center}
\includegraphics[scale=0.45]{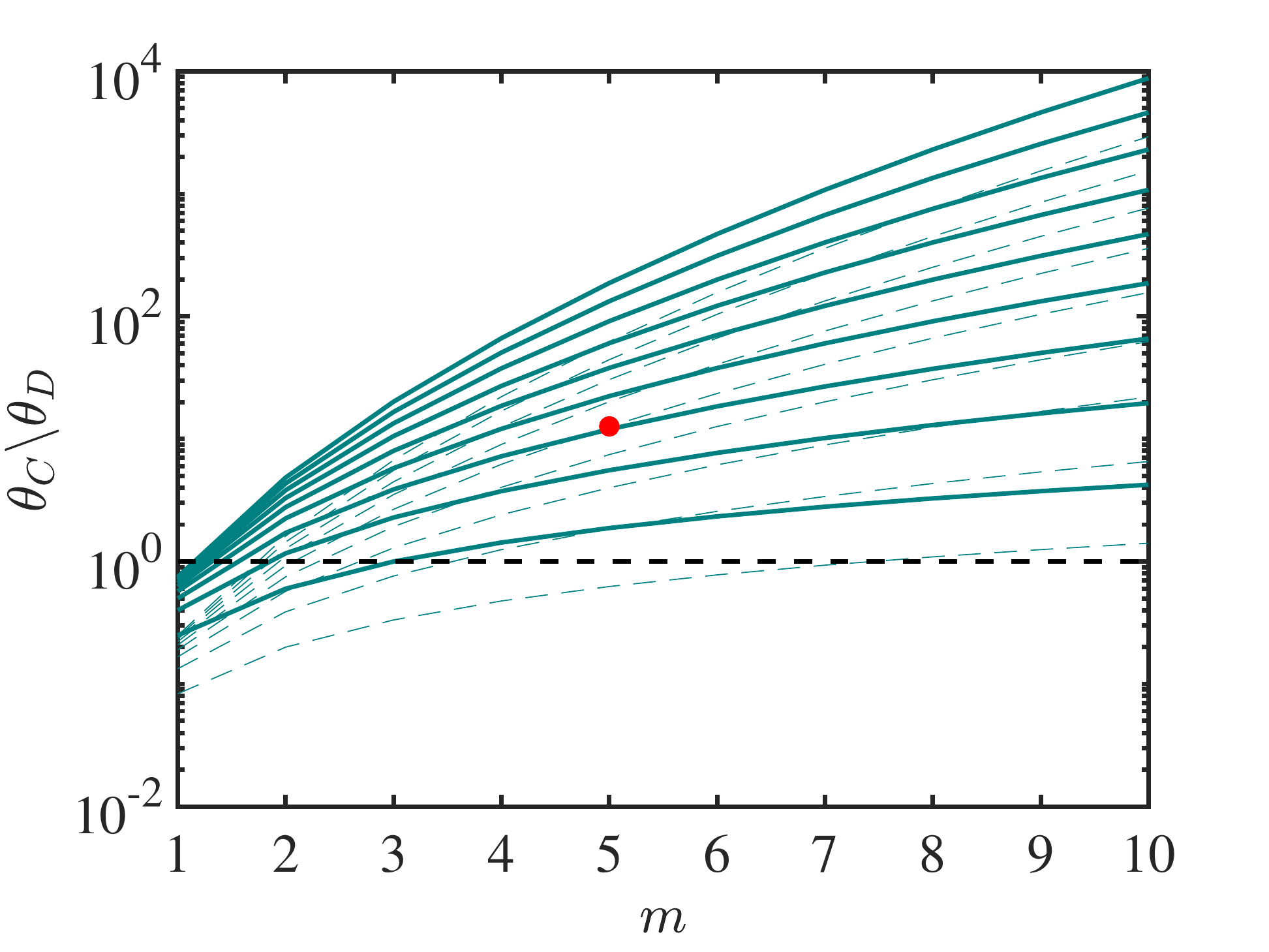}
\caption{Visualisation of the ratio of the number of parameters required for an expression in the classical monomial basis and the required number for the equivalent decoupled structure. Solid lines corresponds to one-branch functions with a single output, i.e. $n=1$, for varying number of inputs $m$ and degrees $d=2,\ldots,10$ (indicated with different lines, larger $d$ results in larger values of $\theta_C\backslash \theta_D$). The dashed lines depict the scenario for three-branch models. The red dot in corresponds to decoupled functions with $n=1$, $m=5$, $d=5$, and $r=3$. Without constant terms, such decoupled functions are described by 36 parameters. Expressed in the classical monomial basis, 456 parameters are used, which is more than 12 times as many.}
\label{f:para_count}
\end{center}
\end{figure}

In many cases decoupled functions are more efficient parameterisations of the nonlinearity. This is often true for polynomials. In the standard monomial basis, the number of parameters grows combinatorially with the number of inputs $m$ and the degree $d$. The parameter-count of a decoupled polynomial form, on the other hand, grows only linearly as a function of $m$ and $d$. The latter is illustrated in Fig.~\ref{f:para_count}. The figure is constructed by considering decoupled polynomial functions with one branch ($r=1$, these are the solid lines) and one output ($n=1$) of varying input size $m$ and degrees $d=2,\ldots,10$. Each such decoupled polynomial can be translated to the classical basis, resulting in a certain number of parameters. The ratio of the numbers obtained for the coupled and the decoupled forms make up the $y$-axis. It illustrates that the decoupled representation is more efficient, irrespective of the degree for $m\ge 2$. Considering more branches ($r\ge1$) shifts the curves downwards, postponing the crossing of the one-to-one ratio indicated by the black dashed line. The scenario for the case of $r=3$ is indicated by dashed curves.

The red dot in Fig.~\ref{f:para_count} corresponds to decoupled functions with the following properties: $n=1$, $m=5$, $d=5$, and $r=3$. Without constant terms, such decoupled functions are described by 36 parameters. Expressed in the classical monomial basis, 456 parameters are used, which is more than 12 times as many. This illustrates that it is often worthwhile to consider a decoupled representation of the nonlinearity. 

To put the algorithm to the test, a decoupled polynomial with random parameters (\rev{drawn from $\mathcal{N}(0,1)$}) of the described shape is generated. From this function the Jacobian tensor is constructed using 200 operating points, drawn from \rev{$\mathcal{U}\left(-1.5,1.5\right)$}. Note that given the equivalency of the coupled and the decoupled polynomial form, the Jacobian tensor constructed from either expression is identical. Next, the implicit algorithm is used to retrieve decoupled forms for a scan over $r$. The default setting using two filters: a 3-points left and a 3-points right filter was selected. The results from a fifth order polynomial parameterisation of the branches are listed in Table \ref{t:2}.
 
\begin{table}[h]
\caption{Relative root-mean-squared error on the function approximation of the polynomial reduction example. Bottom row corresponds to the ratio of numbers of parameters obtained for the coupled and the decoupled functions.}
\label{t:2}
\begin{center}
\renewcommand*{\arraystretch}{1.2}
\begin{tabular}{| c  c | c | c | c | c |}
\cline{3-6}
\multicolumn{2}{c|}{} & $r=1$ & $r=2$ & $r=3$ & $r=4$  \\
\hline
Implicit & $e_{{1}}$ & 8.6 \% & 4.7 \% & 0.60 \% & 0.63 \%  \\ 
\hline
\hline
\multicolumn{2}{|c|}{$\theta_C \backslash \theta_D$} & 38 & 19 & 12.7 & 9.5 \\
\hline
\end{tabular}
\end{center}
\end{table}

It is important to stress that all coupled polynomial functions can be decoupled (see Section \ref{s:approx}). In this example a case was constructed belonging to the subset which require a low number of branches in the decoupled form, leading to significant reductions in the number of parameters. This turns out to be an important class of functions, frequently encountered in practice. The examples in \cite{decuyper2021,decuyper2021NARX,csurcsia2021,csurcsia2022} show that nonlinearities found in dynamical models can often be brought back to a low number of underlying characteristic univariate functions. This can lead to insight into the system's behaviour.

\begin{con}
An important subset of the class of coupled polynomial functions can be represented by a decoupled form containing a low number of univariate polynomial branches. This results in a significant reduction in the number of parameters and may lead to an improved understanding of the relationship.
\end{con}

\section{Case study 2: a neural network example}
\label{s:case_neural}
To illustrate the universal approximation properties of the decoupled form, also a non-polynomial example is considered. We will show that popular feed-forward sigmoidal neural networks may be parameterised more efficiently by single-layer decoupled functions with tailored activation (branch) functions.

A feed-forward neural network has a sequential form in which input features cascade through the layers, ultimately forming the output. Denoting the simulated model output by $\boldsymbol{q} \in \mathbb{R}^{n_L}$ and the input at a single time step by $\boldsymbol{p} \in \mathbb{R}^{n_0}$, we have that
\begin{subequations}
\begin{align}
\boldsymbol{x}^{(0)} &= \boldsymbol{p}, \\
\boldsymbol{x}^{(l)} &= \boldsymbol{f}^{(l)}(\boldsymbol{x}^{(l-1)}), \quad l = 1,\dots, L-1\\
\boldsymbol{q} &= \boldsymbol{f}^{(L)}(\boldsymbol{x}^{(L-1)}),
\end{align}
\end{subequations}
where $L$ denotes the final layer. The applied function in the $l$th layer usually has the following form
\begin{equation}
\boldsymbol{f}^{(l)}(\boldsymbol{x}^{(l-1)}) = \boldsymbol{\sigma}^{(l)}(\boldsymbol{W}^{(l)} \boldsymbol{x}^{(l-1)}+\boldsymbol{b}^{(l)}).
\end{equation}

If we denote the number of neurons in the $l$-th layer by $n_l$, such that $\boldsymbol{x}^{(l)} \in \mathbb{R}^{n_l}$, we have that $\boldsymbol{W}^{(l)} \in \mathbb{R}^{n_l \times n_{l-1}}$ and $\boldsymbol{b}^{(l)} \in \mathbb{R}^{n_l}$. The model parameters $\{ \boldsymbol{W}^{(l)}, \boldsymbol{b}^{(l)} \}_{l=1}^L$ are referred to as the weights $\boldsymbol{W}^{(l)}$ and biases $\boldsymbol{b}^{(l)}$. The nonlinear activation is a vector function $\boldsymbol{\sigma}^{(l)}: \mathbb{R}^{n_l} \mapsto \mathbb{R}^{n_l}$. In classical neural networks, though, all neurons of a single layer are subjected to the same activation. 

In this example we will consider sigmoidal activation, such that $\sigma_i$ is a sigmoidal function. Additionally, $\boldsymbol{f}^{(L)} \coloneqq \textbf{W}^{(L)} \boldsymbol{x}^{(L-1)}+\boldsymbol{b}^{(L)}$, also known as linear activation, which is the appropriate choice for regression problems \cite{goodfellow2016}. An arbitrary neural network is generated by randomly selecting model parameters, i.e.\ $\{ \boldsymbol{W}^{(l)}, \boldsymbol{b}^{(l)} \}_{l=1}^L\sim \mathcal{N}(0,1)$. The generated network has the following dimensions: $n_0 = 2$, $n_1=15$, $n_2 = 10$, $n_3 = 10$, $n_4 = 5$, $n_5 = 2$, resulting in a total number of 382 parameters. Next we apply the fully automated implicit algorithm (Section \ref{ss:implicit}) to retrieve decoupled functions for a range in $r$. Given that the network is random, it may also contain constant terms on the outputs. To account for the constant terms the procedure of Section \ref{s:DC} is applied.

\begin{figure}
\begin{center}
\begin{subfigure}[b]{\textwidth}
\begin{center}
\includegraphics[width=\textwidth]{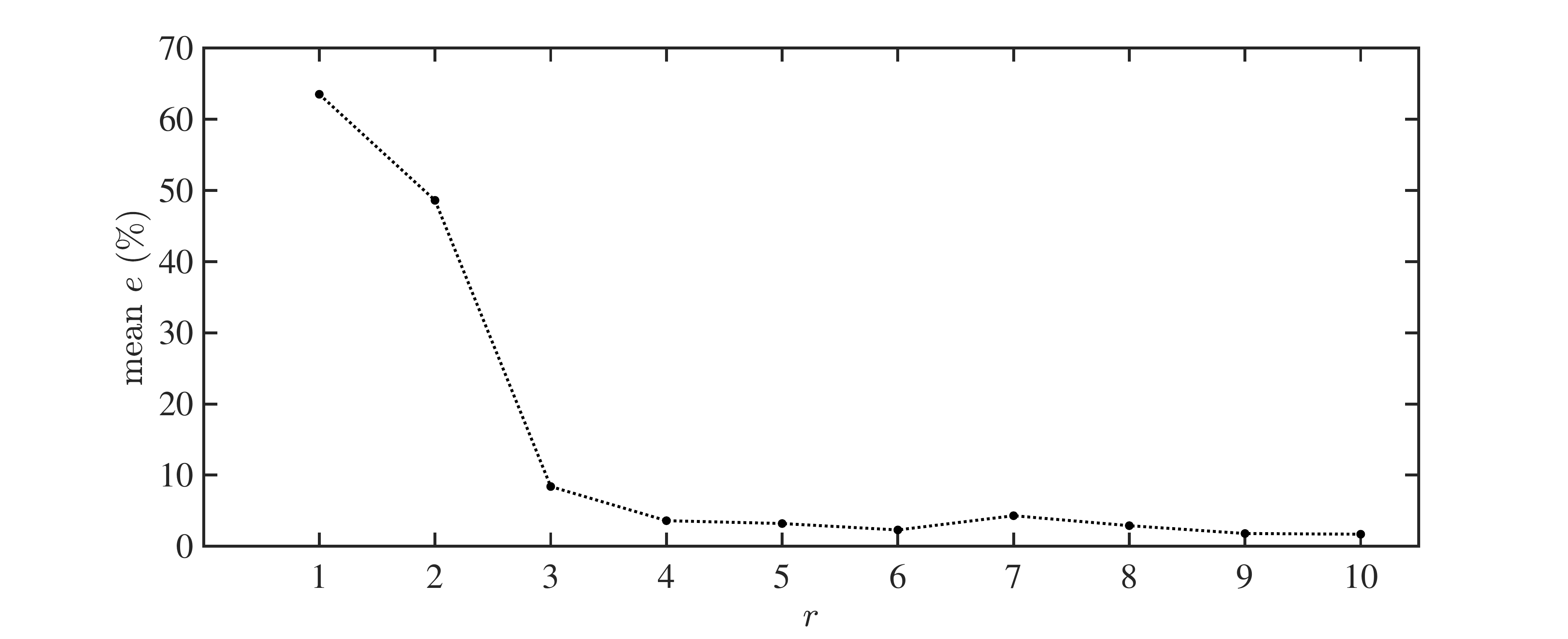}
\caption{}
\end{center}
\end{subfigure}
\begin{subfigure}[b]{\textwidth}
\begin{center}
\includegraphics[width=\textwidth]{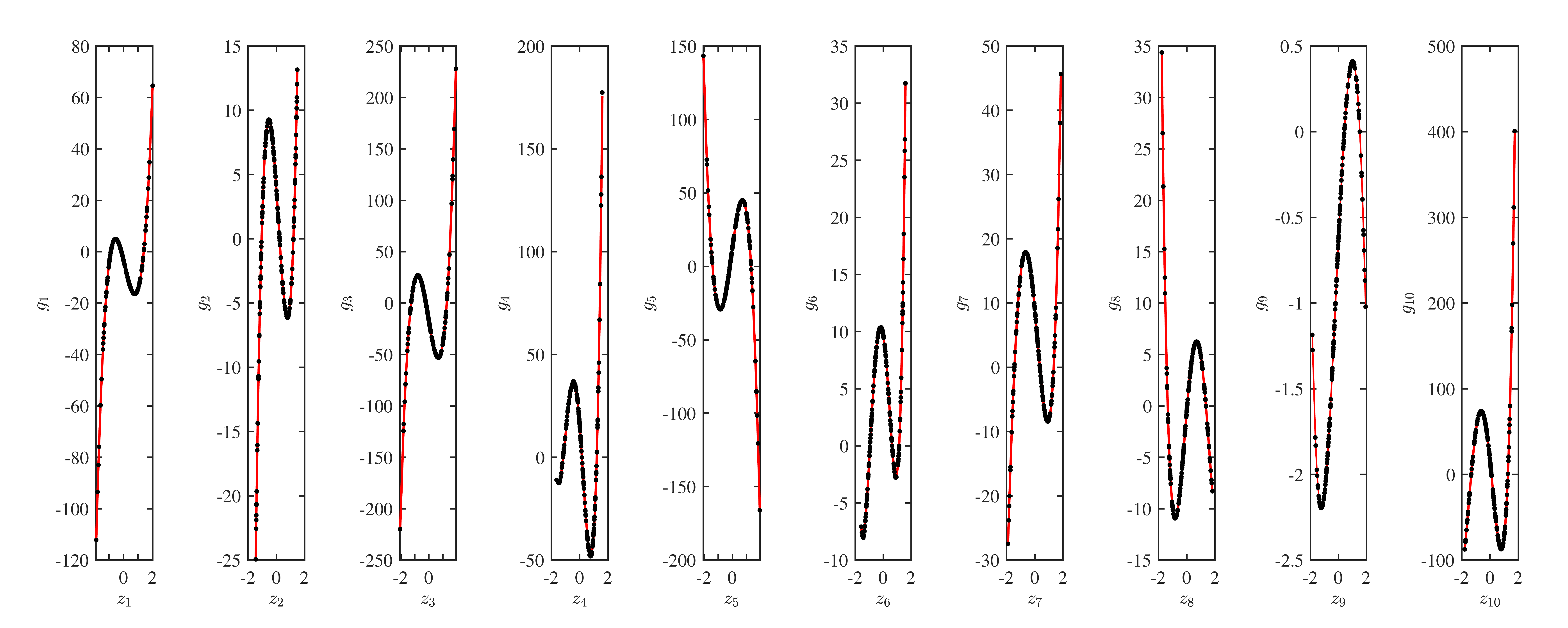}
\caption{}
\end{center}
\end{subfigure}
\caption{(a) Mean function approximation error when decoupling the sigmoidal neural network example. (b) Results for a decoupling using $r=10$ branches. The functions are fitted using a $7^{\text{th}}$ order polynomial.}
\label{f:MLP_graph}
\end{center}
\end{figure}

\rev{It was found that the use of additional filters, e.g.\ a 3-points left, 3-points right, and a 3-points central filter, can be required to facilitate the convergence of the implicit algorithm. This is especially the case for large decoupled functions for which $r> \text{rank}\left(\mathcal{J}\right)$\footnote{The rank of a tensor is here defined as the minimal number of rank-one terms required to minimise \eqref{e:5}.}, which gives rise to an underdetermined tensor factorisations.} The results are presented in Table \ref{t:3}. The mean function approximation error \revv{(following from \eqref{e:rel_er}), computed on the selected operating points,} is additionally visualised in Fig.~\ref{f:MLP_graph}, illustrating that decoupling can be used to balance model complexity to accuracy. Accurate decoupled functions could be retrieved with errors in the order of 1 \% while requiring less than a third of the number of parameters used in the neural network description. A polynomial fit of degree 7 was used in this case. 

\begin{table}[h]
\caption{Relative root-mean-squared error on the function approximation of the sigmoidal neural network example. Bottom row corresponds to the ratio of numbers of parameters obtained for the network and the decoupled functions.}
\label{t:3}
\begin{center}
\renewcommand*{\arraystretch}{1.2} 
\begin{tabular}{| c  c | c | c | c | c | c | c | c | c | c | c|}
\cline{3-12}
\multicolumn{2}{c|}{} & $r=1$ & $r=2$ & $r=3$ & $r=4$ & $r=5$ & $r=6$ & $r=7$ & $r=8$ & $r=9$ & $r=10$  \\
\hline
\multirow{2}{*} & $e_{{1}}$ &  34.2 \% & 9.2  \% & 4.7  \% & 1.6  \% & 1.0 \% & 0.8 \% & 3.7 \% &2.8 \% & 1.1 \% & 1.4 \%  \\ 
& $e_{{2}}$ &  92.8 \% & 88.0 \% & 12.1 \% & 5.6 \%& 5.3 \% &  3.8 \% & 5.0 \% & 3.0 \%& 2.5 \%& 2.1 \% \\
\hline
\hline
\multicolumn{2}{|c|}{mean $e$} & 63.5  \% & 48.6 \% & 8.4 \%  & 3.6 \%  & 3.2 \% & 2.3 \% & 4.3 \% & 2.9 \% & 1.8 \% & 1.7 \% \\
\hline
\hline
\multicolumn{2}{|c|}{$\theta_{NN} \backslash \theta_D$} & 27.3  & 15.3  & 10.6  & 8.1  & 6.6 & 5.5 & 4.8 & 4.2 & 3.7 & 3.4  \\
\hline
\end{tabular}
\end{center}
\end{table}

\begin{con}
Decoupling is a generic tool. The universal approximation theorem states that all continuous real functions may be approximated arbitrary well by a decoupled form. The required complexity of the decoupled function (number of branches) depends on the underlying relationship alone, and not on the family of functions used to express it.  
\end{con}

\begin{con}
Large decoupling problems, for which $r> \text{rank}\left(\mathcal{J}\right)$, give rise to underdetermined tensor factorisation. In such cases additional filters may be required to restrict the solutions of the implicit algorithm to a meaningful subset.
\end{con}

\revv{
\section{Case study 3: NARX models}
\label{s:case_NARX}
An important benefit from function decoupling is that also nonlinear dynamical models, as they appear in system identification, can be handled. Note that nonlinear (data-driven) dynamical models rely on multivariate mapping functions as description of the nonlinearity. This is the case for nonlinear state-space models, where either a multivariate polynomial \cite{decuyper2021PNLSS,paduart2010}, or a neural network is used \cite{schoukensM2021}, but also for NARX models \cite{billings2013}. 

\subsection{Single-output Polynomial NARX models}
In this section we will show that the tools developed in this work allow to decouple NARX models in general, and single-output NARX models in particular. Single-output functions are out of the scope of the original decoupling method of \cite{dreesen2014}. Given that a collection of Jacobians of a single-output function results in a matrix, rather than a third-order tensor, one cannot rely on the uniqueness properties of tensor decompositions, which is the cornerstone of the method of \cite{dreesen2014}. An alternative approach for decoupling single-output NARX models was proposed in \cite{karami2021}. The procedure of \cite{karami2021} resorts to a decomposition of the Hessian, combined with polynomial constraints on one of the factors. The benefits of the tools presented in this work lie in the fact that, both the expensive computation of the Hessian, as well as the polynomial constraints can be avoided. Recall that in this work non-parametric estimates of $g_i$ are obtained, i.e. without having to predefine a function family.

NARX models describe dynamical nonlinear behaviour by relating the current output sample to both past output samples, and current and past input samples. Let
\begin{equation}
\label{e:x}
\boldsymbol{p}(t) \coloneqq \{u(t), u(t-1), \dots, u(t-n_u), y(t-1), \dots, y(t-n_y)\}.
\end{equation}
A general single-output NARX model then assumes the form \citep{billings2013}
\begin{equation}
\label{e:sim}
y(t) = F(\boldsymbol{p}(t))+e(t),
\end{equation}
with $u$ the input, and $y$ the measured output. $F: \mathbb{R}^{n_u + n_y +1} \rightarrow \mathbb{R}$ is a static multiple-input single-output (MISO) nonlinear function and $e(t)$ is an equation error which is assumed to be a sequence of independent identically distributed (IID) random variables. The function $F$ may be described by any function family, e.g.\ a neural network. Often, however, a basis expansion is preferred. In that case, a direct estimate of the model parameters follows from a linear regression problem by using the measured outputs in the regressor, as is done in \eqref{e:x} (minimising the equation error). A popular choice is the polynomial basis, leading to so-called Polynomial-NARX or P-NARX models.

The downside of P-NARX models is that they suffer from the curse of dimensionality. As was discussed in section \ref{ss:poly_reduction}, the number of model parameters grows combinatorially with the number of delayed inputs, outputs, and nonlinear degree, contrary to the linear growth with $r$ and the degree in case of a decoupled polynomial. 

\subsection{Benchmark problem: the forced Duffing oscillator}
\label{s:benchmark}
The decoupling of single-output P-NARX models is demonstrated on a nonlinear benchmark data set of the forced Duffing oscillator. The idealised system equation reads
\begin{equation}
\label{e:diff_SB}
m \ddot{y}(t) + c \dot{y}(t) + k(y(t))y(t) = u(t),
\end{equation}
with mass $m$, viscous damping $c$, and a nonlinear spring $k(y(t))$. The presumed displacement, $y(t)$, is considered the output and the presumed force, $u(t)$, is considered the input. Overdots denote the derivative with respect to time. The system can be interpreted as a cubic hardening spring given the position-dependent stiffness
 \begin{equation}
\label{e:SB_kNL}
k(y(t))=\alpha+\beta y^2(t).
\end{equation}

Data are obtained from an analogue electrical circuitry. 
\textbf{The training set} consists of 9 realisations of a random-phase odd multisine \cite{pintelon2013}. The period of the multisine is $\rfrac{1}{f_0}$ with $f_0= \rfrac{f_s}{8192}$ Hz and $f_s\approx 610$ Hz. The number of excited harmonics is $L=1342$ resulting in an $f_{\text{max}}\approx 200$ Hz. Each multisine realisation is given a unique set of phases that are independent and uniformly distributed in $[0,2\pi[$. The signal to noise ratio at the output is estimated at approximately 40 dB. 
As \textbf{test set}, a filtered Gaussian noise sequence of the same band width and with a linearly increasing amplitude is used. 

The data are part of three benchmark data sets for nonlinear system identification described in \cite{wigren2013} and used in a number of works among which \cite{karami2021,noel2018,ljung2004,sragner2004,schoukens2019}. 

\subsubsection{Reference P-NARX model}
\label{ss:reference_PNARX}
Using the System Identification toolbox in MATLAB a P-NARX model is estimated with the following properties: $n_u = 1$, $n_y = 3$, and $d = 3$. All cross-term monomials where included leading to a model with 55 parameters. The model is estimated in `prediction'-mode, i.e.\ by minimising the equation error in \eqref{e:sim}. The accuracy of the model will, however, be evaluated on the basis of a simulation error, i.e.\ by reinserting the modelled output back into the regressor
\begin{equation}
\boldsymbol{p}_s(t) \coloneqq \{u(t), u(t-1), \dots, u(t-n_u), \hat{y}_s(t-1), \dots, \hat{y}_s(t-n_y)\},
\end{equation}
\begin{equation}
\label{e:pred}
\hat{y}_s(t) = F(\boldsymbol{p}_s(t)),
\end{equation}
where the subscript $s$ refers to the simulation setting. A relative root-mean-square simulation error may then be defined as
\begin{equation}
\label{e:rms}
e_{\text{rms}} = \frac{\sqrt{\frac{1}{N}\sum_{t=1}^N\left(\hat{y}_s(t)-y(t)\right)^2}}{\sqrt{\frac{1}{N}\sum_{t=1}^N\left(y(t)-\mathbb{E}(y(t))\right)^2}} \times 100.
\end{equation}
The estimation process returns an accurate P-NARX model yielding a simulation error $e_{\text{rms}}$ of 1.01\% on the test set.

\subsubsection{Decoupled P-NARX model}
In this case study the use of the explicit smoothness method (section \ref{ss:explicit}) is demonstrated. We will construct decoupled NARX models of the form
\begin{equation}
F(\boldsymbol{p}_s(t)) \approx \boldsymbol{W}\boldsymbol{g}\left(\boldsymbol{V}^{\top}\boldsymbol{p}_s(t)\right).
\end{equation}
Given that the presented method allows to treat $r$ as a design parameter, a range of models of varying complexity can be constructed. We will consider the range $r=1,\ldots,6$. The explicit method additionally requires a scan over the hyperparameter $\lambda$. A coarse search over $\sqrt{\lambda} = 10^{-1},10^0,10^1,10^2,10^3,10^5$, is conducted. The univariate branches are in this case parameterised using third-degree polynomial functions. For every value of $r$, the decoupled model yielding the best performance on the training data (according to \eqref{e:rms}) is selected. The models are compared to the reference P-NARX model on the basis of the test set performance in Fig.~\ref{f:NARX_results}. Observe that decoupled models with $r\ge3$, denoted by `o' markers, result in a performance similar to that of the reference model (indicated by the red line).

If needed, the decoupled model can be subjected to a post-optimisation step. In that case the model parameters are additionally tuned on the basis of the training data given a mean-squared error cost function
\begin{equation}
\label{e:post_opt}
\underset{\boldsymbol{W},\boldsymbol{V},\boldsymbol{\theta}}{\operatorname{arg~min}}~ \sum_{t=1}^{N_T} \left(y(t) - \boldsymbol{W}\boldsymbol{g}\left(\boldsymbol{V}^{\top}\boldsymbol{p}_s(t)\right)\right)^2,
\end{equation}
with $\boldsymbol{\theta}$ storing the polynomial coefficients of $\boldsymbol{g}$, and $N_T$ the length of the training set. The test set performance of the post-optimised models is indicated by `*' in Fig.\ref{f:NARX_results}. Notice that additional tuning on the training set is not guaranteed to improve the test set result.
\begin{figure}
\begin{center}
\includegraphics[scale=0.3]{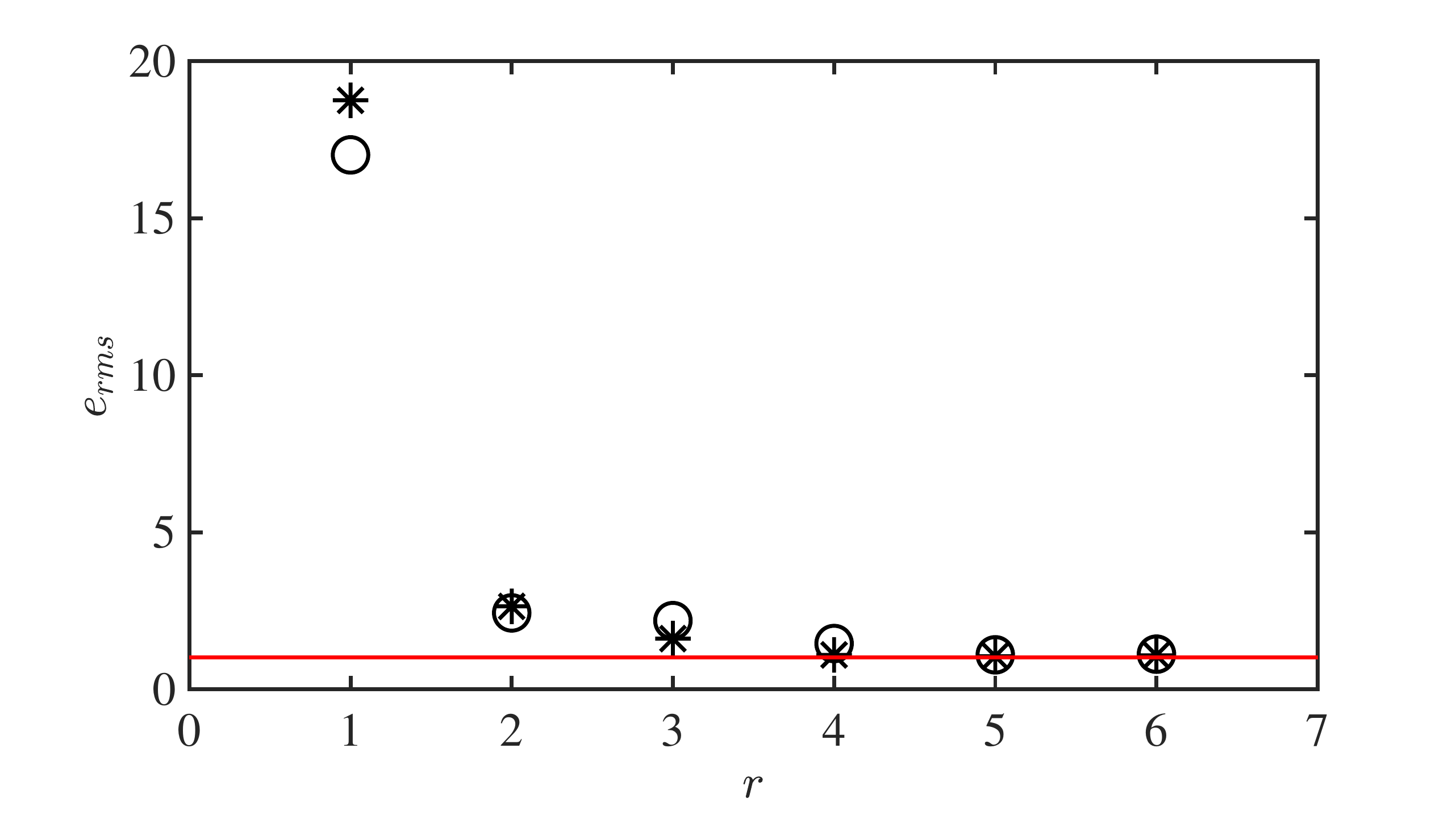}
\caption{\color{black}Test set performance of the decoupled NARX models, obtained using the explicit decoupling algorithm and starting from the reference P-NARX model of section \ref{ss:reference_PNARX}. `o' indicate the performance of the models without post-optimistion, `*' denotes the post-optimised models, found according to \eqref{e:post_opt}. The red line corresponds to the performance of the reference P-NARX model and corresponds to a value of $e_{\text{rms}} = 1.01\%$.}
\label{f:NARX_results}
\end{center}
\end{figure}

The benefit of obtaining a decoupled NARX model is twofold:
\begin{enumerate}
\item The decoupled model with $r=3$ requires only 30 parameters (compared to 55 for the reference model), while resulting in a similar performance. The complexity of the model can, moreover, easily be balanced to the accuracy. For a given application the accuracy of the $r=2$ model might still be sufficient, resulting in an even larger model reduction. 
\item Given that the nonlinear elements are now described by univariate nonlinearities, they may easily be visualised, as is done in Fig.~\ref{f:NARX_r3}. The functions reveal a dominantly cubic relationship in branch 1 and 3, and a dominantly linear form in 2. This relates to the physical model \eqref{e:diff_SB}, containing both linear and cubic stiffness.
\end{enumerate}

\begin{figure}
\begin{center}
\includegraphics[scale=0.3]{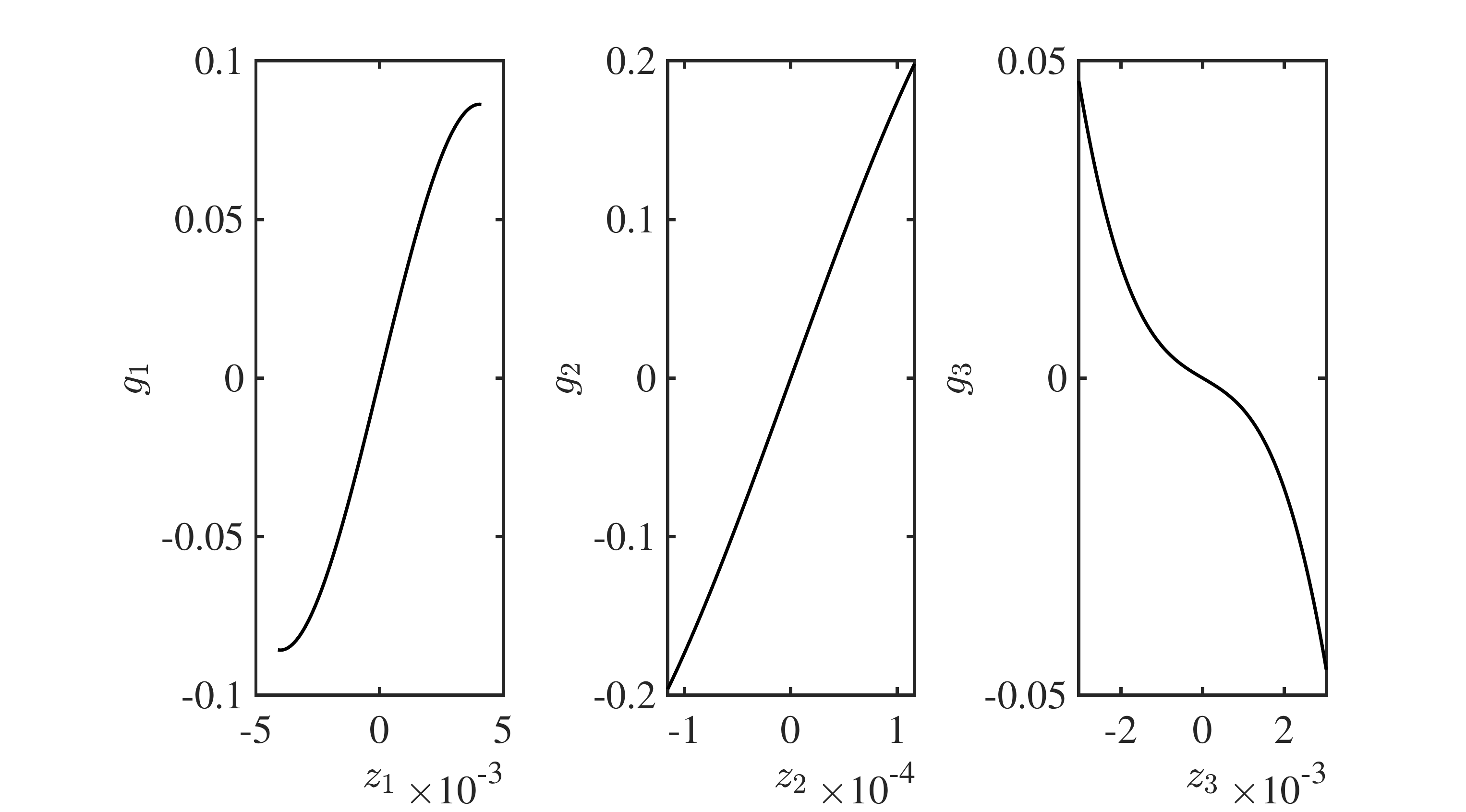}
\caption{\color{black} Nonlinear branch functions of the post-optimised $r=3$ decoupled NARX model. The functions reveal a dominantly cubic relationship in branch 1 and 3, and a dominantly linear form in 2. This relates to the physical model \eqref{e:diff_SB}, containing both linear and cubic stiffness.
}
\label{f:NARX_r3}
\end{center}
\end{figure}

A similar test set performance was obtained for the competing method of \cite{karami2021}, yielding a value of 1.78\% given the same settings, i.e.\ $r=4$, $n_u=1$ and $n_y=3$ \cite{decuyper2021NARX}, and a value of 0.9\% in case of a slightly larger model with $r=4$, $n_u=2$ and $n_y=3$ \cite{karami2021}.

\begin{con}
Decoupling can be used to process dynamical models. The result is a reduced model with tractable nonlinear elements. A visualisation of the nonlinear functions can provide insight into the underlying physical system.
\end{con}
\begin{con}
The filtered decoupling procedures, presented in this work, allow to decoupled single-output functions, which was out of the scope of the original method of \cite{dreesen2014}. As a result, also single-output NARX models can be handled.
\end{con}
}

\section{Discussion}
\label{s:discussion}

The examples presented in this work illustrate that a decoupled function can be a useful alternative representation of the classical basis expansion or neural network. It is, however, important to note that decoupling techniques are designed to be a subsequent step in the modelling process. The decoupled form is distilled from a \emph{coupled} function. The sequence of the procedure is imperative for the following reasons:
\begin{itemize}
\item Direct estimation of a decoupled function would involve solving a hard nonlinear optimisation problem (output error minimisation). Moreover, a parameterisation of the branches would be required from the start, fixing the function family. Without proper initialisation, the practical use of decoupled forms would be limited \revv{to small functions, i.e.\ with few branches}.
\item Classical regression tools are powerful because they rely on well established parameter estimation procedures. Basis expansions result in linear problems, allowing solutions to be found in one step. Neural networks exploit the topology of high dimensional cost functions. It can be shown that it becomes exponentially unlikely for the eigenvalues of the Hessian at a critical point to become all positive (or negative). Therefore most critical points are saddle points, rather than local minima \cite{dauphin2014}. Moreover, if you do get trapped at a saddle point,  the analysis seems to show that the large number of saddle points have very similar values of the objective function, explaining why neural networks are not very sensitive to initialisation \cite{lecun2015}.
\end{itemize}

It is therefore proposed to exploit the favourable optimisation properties of the classical methods and promote the use of decoupled forms only in a second step. Having obtained a decoupled function, it may be beneficial to consider it as an initialisation and subject it to further tuning using nonlinear optimisation. Such a post-optimisation step, based on input-output data, is useful since it may remove any bias which was present in the originally obtained coupled function.

\section{Conclusion}

Decoupling multivariate functions is a generic tool which may be applied to all function families. In many cases a more efficient parameterisation of the nonlinear relationship is retrieved. Moreover, the structured decoupled form may reveal a set of characteristic functions which lie at the heart of the nonlinear mapping, potentially leading to insight. Two algorithms based on filtered tensor decompositions of first order derivative information were introduced. The methods are nonparametric, allowing decoupled functions of any type to emerge. A parameterised form is only obtained in a final isolated step. Decoupling provides a means for model reduction, directly applicable to i.a.\ nonlinear system identification and machine learning.

\bibliography{entirebib}

\appendix
\section{Non-equidistant finite differences}    
\label{a:1}
A finite difference expression for non-equidistantly spaced points can be obtained through Lagrange interpolation \citep{singh2009}. Let $\boldsymbol{z}_s \in \mathbb{R}^N$ be a vector of $z$-values, sorted in ascending order and let $\boldsymbol{g}_s \in \mathbb{R}^N$ be the vector of corresponding evaluations of the function $g(z)$, sorted according to $z$. Consider $\boldsymbol{z}_w$ to be a window of $k$ neighbouring elements such that $\boldsymbol{z}_w \subset \boldsymbol{z}_s$ and let $\boldsymbol{g}_w$ be the corresponding window of function evaluations such that $\boldsymbol{g}_w \subset \boldsymbol{g}_s$. The Lagrange polynomial is the polynomial of lowest degree which interpolates all $k$ points within the window \citep{lagrange1812,suli2003}. It is a linear combination of basis functions
\begin{equation}
L(z) \coloneqq \sum_{j=1}^k \boldsymbol{g}_w[j]l_j(z),
\end{equation}
where $l_j$ are known as Lagrange basis polynomials
\begin{equation}
l_j(z) \coloneqq \prod_{\begin{matrix} 1 \le i \le k \\ i \ne j \end{matrix}} \frac{z - \boldsymbol{z}_w[i]}{\boldsymbol{z}_w[j]-\boldsymbol{z}_w[i]}.
\end{equation}
At the grid points $\boldsymbol{z}_w$, the basis polynomials have the following two properties:
\begin{enumerate}
\item For all $s \ne j$ the product will be zero since the numerator $\boldsymbol{z}_w[s] - \boldsymbol{z}_w[i]$ attains zero at $s = i$.
\begin{equation}
\forall (s \ne j) : l_j(\boldsymbol{z}_w[s]) = \prod_{\begin{matrix} 1 \le i \le k \\ i \ne j \end{matrix}} \frac{\boldsymbol{z}_w[s] - \boldsymbol{z}_w[i]}{\boldsymbol{z}_w[j]-\boldsymbol{z}_w[i]} = 0. 
\end{equation}
\item On the other hand
\begin{equation}
l_j(\boldsymbol{z}_w[j]) = \prod_{\begin{matrix} 1 \le i \le k \\ i \ne j \end{matrix}} \frac{\boldsymbol{z}_w[j] - \boldsymbol{z}_w[i]}{\boldsymbol{z}_w[j]-\boldsymbol{z}_w[i]} = 1.
\end{equation}
\end{enumerate}
Given that all basis polynomials are zero when ${s \ne j}$, and since they equal to one for $s =j$, it follows that $L(\boldsymbol{z}_w[j]) = \boldsymbol{g}_w[j]$, meaning that $L(z)$ passes exactly through all $k$ points in the window. 

Having found a function which relates neighbouring points, also a derivative relationship between neighbouring points (i.e.\ a finite difference) may be obtained. A first order derivative expression is given by
\begin{equation}
\label{e:L1}
L^{(1)}(z) \coloneqq \sum_{j=1}^k \boldsymbol{g}_w[j]l^{(1)}_j(z), 
\end{equation}
which is a weighted sum with the weights following from $l^{(1)}_j(z) \coloneqq \frac{dl_j(z)}{dz}$
\begin{equation}
l^{(1)}_j(z) \coloneqq \sum_{\begin{matrix} s = 1 \\ s \ne j \end{matrix}}^k \left[ \frac{1}{\boldsymbol{z}_w[j] - \boldsymbol{z}_w[s]} \prod_{\begin{matrix} i =1 \\ i \ne (s,j) \end{matrix}}^k \frac{z - \boldsymbol{z}_w[i]}{\boldsymbol{z}_w[j] - \boldsymbol{z}_w[i]}  \right].
\end{equation}
\begin{figure}
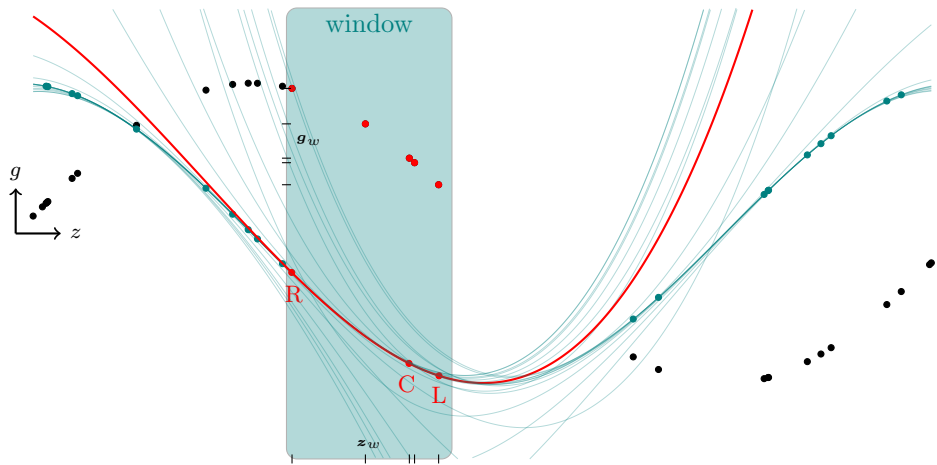

\begin{center}
\include{Figures/Lagrange_illustration}
\caption{Illustration of the use of Lagrange polynomials to compute non-equidistant finite differences. Black markers indicate evaluations of a sine function on a non-equidistant grid of points. Based on a sliding window of 5 points the Lagrange polynomial is constructed. The derivative of this polynomial, i.e.\ $L^{(1)}(z)$ is indicated by a thin solid line (for every window). Evaluating $L^{(1)}(z)$ at the grid points results in a finite difference approximation. $L^{(1)}(z)$ corresponding to the indicated window is highlighted in red. If this curve is used to estimate the derivative in the most forward point of the window a left (or forward) finite difference is obtained. Analogously a central and a right (or backward) finite difference may be obtained. Collecting, for example, the central difference estimate of every window reveals the cosine function indicated by teal markers.}
\label{f:Lagrange}
\end{center}
\end{figure}
The first order derivative based on Lagrange polynomials, i.e.\ $L^{(1)}(z)$, is illustrated on a sampled sine function in Fig.~\ref{f:Lagrange}.

Notice that the linear combination, expressed by \eqref{e:L1}, corresponds to the operations encoded in the rows of the matrix $\boldsymbol{D}$ (\eqref{e:D}). When computing a 3-points finite difference, $k=3$, the weights $l^{(1)}_j(z)$ with $j=1,\ldots,k$ form the elements of $\boldsymbol{D}$. Contrary to \eqref{e:D}, all weights may differ from one another since they are computed on the basis of the non-equidistant spacing of the $k$ points in $\boldsymbol{z}_w$. Going from one row to the next may be seen as sliding the window of the finite difference operator over the $z$-axis. The positioning of the window allows to switch from forward to central to backward differencing, additionally also the length may be changed. This flexibility enables to compute a number of alternative finite difference filters, which are all approximations of the first order derivative. Consider the following examples:
\begin{itemize}
\item \textbf{3-points central finite difference filter}

The $i$th row of a 3-points ($k=3$) central differencing filter of size $N$, $\boldsymbol{D}_C$, is based on the window $\boldsymbol{z}_{w_i} = \left[ \boldsymbol{z}_s[i-1] \quad \boldsymbol{z}_s[i] \quad \boldsymbol{z}_s[i+1] \right]^{\top}$. The weights are accordingly found by evaluating the basis functions $l_j^{(1)}(z)$ in the central point of the window, i.e.\ $l_j^{(1)}(\boldsymbol{z}_{w_i}[2])$ with $j=1,\ldots,k$. The following filter is obtained:

\renewcommand{\arraystretch}{3}

\begin{equation}
\label{e:D1}
\resizebox{0.85\textwidth}{!}{$
\boldsymbol{D}_C \coloneqq  \left[ \begin{array}{cccccccc}
 l_1^{(1)}(\boldsymbol{z}_{w_1}[1]) &  l_2^{(1)}(\boldsymbol{z}_{w_1}[1]) & l_3^{(1)}(\boldsymbol{z}_{w_1}[1]) & 0 & \dots &  0 \\
  l_1^{(1)}(\boldsymbol{z}_{w_2}[2]) &  l_2^{(1)}(\boldsymbol{z}_{w_2}[2]) & l_3^{(1)}(\boldsymbol{z}_{w_2}[2]) & 0 & \dots &  0 \\
   0 &  l_1^{(1)}(\boldsymbol{z}_{w_3}[2]) &  l_2^{(1)}(\boldsymbol{z}_{w_3}[2]) & l_3^{(1)}(\boldsymbol{z}_{w_3}[2]) & \ddots & \vdots \\
 \vdots & \ddots & \ddots & \ddots & \ddots & 0  \\
  0 & \dots & 0 & l_1^{(1)}(\boldsymbol{z}_{w_{N-1}}[2]) &  l_2^{(1)}(\boldsymbol{z}_{w_{N-1}}[2]) & l_3^{(1)}(\boldsymbol{z}_{w_{N-1}}[2])\\
    0 & \dots & 0 & l_1^{(1)}(\boldsymbol{z}_{w_{N}}[3]) &  l_2^{(1)}(\boldsymbol{z}_{w_{N}}[3]) & l_3^{(1)}(\boldsymbol{z}_{w_{N}}[3])\\
 \end{array} \right]. $}
\end{equation}
The first and the last row face a boundary and are treated as 3-points right (backward) and 3-points left (forward) filters, correspondingly.
\revv{For convenience, an explicit expression is provided for the 3 non-zero entries of th $i$th row
\begin{equation}
l_1^{(1)}\left(\boldsymbol{z}_{w_i}[2]\right) = \frac{\boldsymbol{z}_{w_i}[2]-\boldsymbol{z}_{w_i}[3]}{\left(\boldsymbol{z}_{w_i}[1]-\boldsymbol{z}_{w_i}[2]\right)\left(\boldsymbol{z}_{w_i}[1]-\boldsymbol{z}_{w_i}[3]\right)},
\end{equation}
\begin{equation}
l_2^{(1)}\left(\boldsymbol{z}_{w_i}[2]\right) = \frac{\boldsymbol{z}_{w_i}[1]-2\boldsymbol{z}_{w_i}[2]+\boldsymbol{z}_{w_i}[3]}{\left(\boldsymbol{z}_{w_i}[1]-\boldsymbol{z}_{w_i}[2]\right)\left(\boldsymbol{z}_{w_i}[2]-\boldsymbol{z}_{w_i}[3]\right)},
\end{equation}
\begin{equation}
l_3^{(1)}\left(\boldsymbol{z}_{w_i}[2]\right) = \frac{\boldsymbol{z}_{w_i}[2]-\boldsymbol{z}_{w_i}[1]}{\left(\boldsymbol{z}_{w_i}[3]-\boldsymbol{z}_{w_i}[2]\right)\left(\boldsymbol{z}_{w_i}[3]-\boldsymbol{z}_{w_i}[1]\right)}.
\end{equation}
}
\item \textbf{3-points left finite difference filter}

The $i$th row of a 3-points ($k=3$) left differencing filter of size $N$, $\boldsymbol{D}_L$, is based on the window $\boldsymbol{z}_{w_i} = \left[ \boldsymbol{z}_s[i-2] \quad \boldsymbol{z}_s[i-1] \quad \boldsymbol{z}_s[i] \right]^{\top}$. The weights are accordingly found by evaluating the basis functions $l_j^{(1)}(z)$ in the most forward point of the window, i.e.\ $l_j^{(1)}(\boldsymbol{z}_{w_i}[3])$ with $j=1,\ldots,k$. The following filter is obtained:

\begin{equation}
\label{e:D2}
\resizebox{0.85\textwidth}{!}{$
\boldsymbol{D}_L \coloneqq  \left[ \begin{array}{cccccccc}
 l_1^{(1)}(\boldsymbol{z}_{w_1}[1]) &  l_2^{(1)}(\boldsymbol{z}_{w_1}[1]) & l_3^{(1)}(\boldsymbol{z}_{w_1}[1]) & 0 & \dots &  0 \\
  l_1^{(1)}(\boldsymbol{z}_{w_2}[2]) &  l_2^{(1)}(\boldsymbol{z}_{w_2}[2]) & l_3^{(1)}(\boldsymbol{z}_{w_2}[2]) & 0 & \dots &  0 \\
   l_1^{(1)}(\boldsymbol{z}_{w_3}[3]) &  l_2^{(1)}(\boldsymbol{z}_{w_3}[3]) & l_3^{(1)}(\boldsymbol{z}_{w_3}[3]) & 0 & \dots & 0 \\
   0 &   l_1^{(1)}(\boldsymbol{z}_{w_4}[3]) &  l_2^{(1)}(\boldsymbol{z}_{w_4}[3]) & l_3^{(1)}(\boldsymbol{z}_{w_4}[3]) & \ddots  & \vdots \\
 \vdots & \ddots & \ddots & \ddots & \ddots & 0  \\
    0 & \dots & 0 & l_1^{(1)}(\boldsymbol{z}_{w_{N}}[3]) &  l_2^{(1)}(\boldsymbol{z}_{w_{N}}[3]) & l_3^{(1)}(\boldsymbol{z}_{w_{N}}[3])\\
 \end{array} \right]. $}
\end{equation}
Given the boundary, the first row is treated as a backward scheme, the second as a central scheme and starting from the third a forward (or left finite difference) is obtained. \revv{Also here, an explicit expression is provided for the 3 non-zero entries of th $i$th row
\begin{equation}
l_1^{(1)}\left(\boldsymbol{z}_{w_i}[3]\right) = \frac{\boldsymbol{z}_{w_i}[3]-\boldsymbol{z}_{w_i}[2]}{\left(\boldsymbol{z}_{w_i}[1]-\boldsymbol{z}_{w_i}[3]\right)\left(\boldsymbol{z}_{w_i}[1]-\boldsymbol{z}_{w_i}[2]\right)},
\end{equation}
\begin{equation}
l_2^{(1)}\left(\boldsymbol{z}_{w_i}[3]\right) = \frac{\boldsymbol{z}_{w_i}[3]-\boldsymbol{z}_{w_i}[1]}{\left(\boldsymbol{z}_{w_i}[2]-\boldsymbol{z}_{w_i}[3]\right)\left(\boldsymbol{z}_{w_i}[2]-\boldsymbol{z}_{w_i}[1]\right)},
\end{equation}
\begin{equation}
l_3^{(1)}\left(\boldsymbol{z}_{w_i}[3]\right) = \frac{\boldsymbol{z}_{w_i}[1]-2\boldsymbol{z}_{w_i}[3]+\boldsymbol{z}_{w_i}[2]}{\left(\boldsymbol{z}_{w_i}[1]-\boldsymbol{z}_{w_i}[3]\right)\left(\boldsymbol{z}_{w_i}[3]-\boldsymbol{z}_{w_i}[2]\right)}.
\end{equation}
}

\item \textbf{3-points right finite difference filter}

The $i$th row of a 3-points ($k=3$) right differencing filter of size $N$, $\boldsymbol{D}_R$, is based on the window $\boldsymbol{z}_{w_i} = \left[ \boldsymbol{z}_s[i] \quad \boldsymbol{z}_s[i+1] \quad \boldsymbol{z}_s[i+2] \right]^{\top}$. The weights are accordingly found by evaluating the basis functions $l_j^{(1)}(z)$ in the backward point of the window, i.e.\ $l_j^{(1)}(\boldsymbol{z}_{w_i}[1])$ with $j=1,\ldots,k$. The following filter is obtained:

\begin{equation}
\label{e:D2}
\resizebox{0.85\textwidth}{!}{$
\boldsymbol{D}_R \coloneqq  \left[ \begin{array}{cccccccc}
 l_1^{(1)}(\boldsymbol{z}_{w_1}[1]) &  l_2^{(1)}(\boldsymbol{z}_{w_1}[1]) & l_3^{(1)}(\boldsymbol{z}_{w_1}[1]) & 0 & \dots &  0 \\
  0 & \ddots & \ddots & \ddots & \ddots & \vdots \\
      \vdots & \ddots &  l_1^{(1)}(\boldsymbol{z}_{w_{N-3}}[1]) &  l_2^{(1)}(\boldsymbol{z}_{w_{N-3}}[1]) & l_3^{(1)}(\boldsymbol{z}_{w_{N-3}}[1]) & 0\\

      0 & \dots & 0 & l_1^{(1)}(\boldsymbol{z}_{w_{N-2}}[1]) &  l_2^{(1)}(\boldsymbol{z}_{w_{N-2}}[1]) & l_3^{(1)}(\boldsymbol{z}_{w_{N-2}}[1])\\
     0 & \dots & 0 & l_1^{(1)}(\boldsymbol{z}_{w_{N-1}}[2]) &  l_2^{(1)}(\boldsymbol{z}_{w_{N-1}}[2]) & l_3^{(1)}(\boldsymbol{z}_{w_{N-1}}[2])\\
    0 & \dots & 0 & l_1^{(1)}(\boldsymbol{z}_{w_{N}}[3]) &  l_2^{(1)}(\boldsymbol{z}_{w_{N}}[3]) & l_3^{(1)}(\boldsymbol{z}_{w_{N}}[3])\\
 \end{array} \right]. $}
\end{equation}
Given the boundary, the last row is treated as a forward scheme, the second to last as a central scheme and starting from the third to last a backward (or right finite difference) is obtained. \revv{An explicit expression is provided for the 3 non-zero entries of th $i$th row
\begin{equation}
l_1^{(1)}\left(\boldsymbol{z}_{w_i}[1]\right) = \frac{-\boldsymbol{z}_{w_i}[2]+2\boldsymbol{z}_{w_i}[1]-\boldsymbol{z}_{w_i}[3]}{\left(\boldsymbol{z}_{w_i}[1]-\boldsymbol{z}_{w_i}[2]\right)\left(\boldsymbol{z}_{w_i}[1]-\boldsymbol{z}_{w_i}[3]\right)},
\end{equation}
\begin{equation}
l_2^{(1)}\left(\boldsymbol{z}_{w_i}[1]\right) = \frac{\boldsymbol{z}_{w_i}[1]-\boldsymbol{z}_{w_i}[3]}{\left(\boldsymbol{z}_{w_i}[2]-\boldsymbol{z}_{w_i}[1]\right)\left(\boldsymbol{z}_{w_i}[2]-\boldsymbol{z}_{w_i}[3]\right)},
\end{equation}
\begin{equation}
l_3^{(1)}\left(\boldsymbol{z}_{w_i}[1]\right) = \frac{\boldsymbol{z}_{w_i}[1]-\boldsymbol{z}_{w_i}[2]}{\left(\boldsymbol{z}_{w_i}[3]-\boldsymbol{z}_{w_i}[1]\right)\left(\boldsymbol{z}_{w_i}[3]-\boldsymbol{z}_{w_i}[2]\right)}.
\end{equation}
}
\end{itemize}
\section{Update formulas of $\boldsymbol{G}$}
\label{a:B}

\subsection{An implicit smoothness objective}
\label{a:B2}

\textbf{Update $\boldsymbol{W} \rightarrow \boldsymbol{W}^+$}

The factor $\boldsymbol{W}$ is to be updated from the objective
\begin{equation} \tag{\ref{e:Wimp}}
\underset{\boldsymbol{W}}{\operatorname{arg~min}}~\Vert \boldsymbol{J}_{(1)} - \boldsymbol{W}((\mathcal{F}_1(\boldsymbol{V}) \circ \boldsymbol{G}) \kr \boldsymbol{V})^{\top} \Vert_F^2
+ \ldots + \Vert \boldsymbol{J}_{(1)} - \boldsymbol{W}((\mathcal{F}_s(\boldsymbol{V}) \circ \boldsymbol{G}) \kr \boldsymbol{V})^{\top} \Vert_F^2.
\end{equation}
Regrouping the norms leads to the equivalent expression
\begin{equation}
\underset{\boldsymbol{W}}{\operatorname{arg~min}}~\left\| \begin{bmatrix} \boldsymbol{J}_{(1)}^{\top} \\ \vdots \\  \boldsymbol{J}_{(1)}^{\top} \end{bmatrix} - \begin{bmatrix} (\mathcal{F}_1(\boldsymbol{V}) \circ \boldsymbol{G}) \kr \boldsymbol{V} \\ \vdots \\ (\mathcal{F}_s(\boldsymbol{V}) \circ \boldsymbol{G}) \kr \boldsymbol{V} \end{bmatrix} \boldsymbol{W}^{\top}\right\|_F^2.
\end{equation}
An analytical update formula is then obtained by defining
\begin{equation}
\boldsymbol{K} \coloneqq \begin{bmatrix} (\mathcal{F}_1(\boldsymbol{V}) \circ \boldsymbol{G}) \kr \boldsymbol{V} \\ \vdots \\ (\mathcal{F}_s(\boldsymbol{V}) \circ \boldsymbol{G}) \kr \boldsymbol{V} \end{bmatrix},
\end{equation}
leading to the least-squares solution
\begin{equation}
\boldsymbol{W}^{+^{\top}} = \left(\boldsymbol{K}^{\top}\boldsymbol{K}\right)^{-1}\boldsymbol{K}^{\top}  \begin{bmatrix} \boldsymbol{J}_{(1)}^{\top} \\ \vdots \\  \boldsymbol{J}_{(1)}^{\top} \end{bmatrix},
\end{equation}
which is solved in a numerically more stable way using the pseudoinverse
\begin{equation}
\boldsymbol{W}^{+^{\top}} = \boldsymbol{K}^{\dagger}  \begin{bmatrix} \boldsymbol{J}_{(1)}^{\top} \\ \vdots \\  \boldsymbol{J}_{(1)}^{\top} \end{bmatrix}.
\end{equation}

\textbf{Update $\boldsymbol{G} \rightarrow \boldsymbol{G}^+$}

\begin{lem} Let \eqref{e:Gimp} be the objective function from which $\boldsymbol{G}$ is to be updated. The claim is that $\boldsymbol{G}$ appears linearly in the objective such that an analytical update formula may be derived.

\begin{equation} \tag{\ref{e:Gimp}}
\underset{\boldsymbol{G}}{\operatorname{arg~min}}~\Vert \boldsymbol{J}_{(3)} - (\mathcal{F}_1(\boldsymbol{V}) \circ \boldsymbol{G})(\boldsymbol{V} \kr \boldsymbol{W})^{\top} \Vert_F^2
+ \ldots + \Vert \boldsymbol{J}_{(3)} -  (\mathcal{F}_s(\boldsymbol{V}) \circ \boldsymbol{G})(\boldsymbol{V}  \kr \boldsymbol{W})^{\top} \Vert_F^2
\end{equation}
\end{lem}

\begin{pf}
Consider the vectorised form of the objective function. For clarity the dependency on $\boldsymbol{V}$ is dropped, i.e.\ $\mathcal{F}_i(\boldsymbol{V}) \coloneqq \mathcal{F}_i$.
\begin{equation}
\underset{\boldsymbol{G}}{\operatorname{arg~min}}~\Vert \text{vec}\left(\boldsymbol{J}_{(3)}\right) - \text{vec}\left((\mathcal{F}_1 \circ \boldsymbol{G})(\boldsymbol{V} \kr \boldsymbol{W})^{\top}\right) \Vert_F^2
+ \ldots + \Vert \text{vec}\left(\boldsymbol{J}_{(3)}\right) -  \text{vec}\left((\mathcal{F}_s \circ \boldsymbol{G})(\boldsymbol{V}  \kr \boldsymbol{W})^{\top}\right) \Vert_F^2
\end{equation}
Denoting the Knonecker product by `$\kron$' we may use the property, $\text{vec}\left(\boldsymbol{A}\boldsymbol{X}\boldsymbol{B}\right) = (\boldsymbol{B}^{\top} \kron \boldsymbol{A})\text{vec}(\boldsymbol{X})$. With $\boldsymbol{A} = \boldsymbol{I}_N$, $\boldsymbol{B} = \left(\boldsymbol{V} \kr \boldsymbol{W} \right)^{\top}$ and $\boldsymbol{X} = \left(\mathcal{F}_i \circ \boldsymbol{G} \right)$ we have that
\begin{equation}
\label{e:AXB}
\operatorname{vec}\left(\left(\mathcal{F}_i \circ \boldsymbol{G}\right) \left(\boldsymbol{V} \kr \boldsymbol{W} \right)^{\top} \right) =
\left( \left(\boldsymbol{V} \kr \boldsymbol{W}\right) \kron \boldsymbol{I}_N \right) \text{vec}\left(\mathcal{F}_i \circ \boldsymbol{G} \right).
\end{equation}
Regrouping the norms and applying the property of \eqref{e:vec}, the objective is rewritten into 
\begin{equation}
\underset{\boldsymbol{G}}{\operatorname{arg~min}}~\left\| \begin{bmatrix} \text{vec}\left(\boldsymbol{J}_{(3)}\right)\\ \vdots \\ \text{vec}\left(\boldsymbol{J}_{(3)}\right) \end{bmatrix}
-\begin{bmatrix}\left( \left(\boldsymbol{V} \kr \boldsymbol{W}\right) \kron \boldsymbol{I}_N \right) \text{blkdiag}\left(\mathcal{F}_1\right)\\ \vdots \\ \left( \left(\boldsymbol{V} \kr \boldsymbol{W}\right) \kron \boldsymbol{I}_N \right) \text{blkdiag}\left(\mathcal{F}_s \right) \end{bmatrix} \text{vec}\left(\boldsymbol{G}\right) \right\|_F^2,
\end{equation}
from which it is clear that $\operatorname{vec}\left(\boldsymbol{G}\right)$ appears linearly in the objective. \qed
\end{pf}

An analytical update formula is obtained by defining
\begin{equation}
\label{e:K}
\boldsymbol{K} \coloneqq \begin{bmatrix}\left( \left(\boldsymbol{V} \kr \boldsymbol{W}\right) \kron \boldsymbol{I}_N \right) \text{blkdiag}\left(\mathcal{F}_1\right)\\ \vdots \\ \left( \left(\boldsymbol{V} \kr \boldsymbol{W}\right) \kron \boldsymbol{I}_N \right) \text{blkdiag}\left(\mathcal{F}_s \right) \end{bmatrix},
\end{equation}
leading to the least-squares solution
\begin{equation}
\text{vec}\left(\boldsymbol{G}^{+}\right) = \left(\boldsymbol{K}^{\top}\boldsymbol{K}\right)^{-1}\boldsymbol{K}^{\top}  \begin{bmatrix} \text{vec}\left(\boldsymbol{J}_{(3)}\right) \\ \vdots \\  \text{vec}\left(\boldsymbol{J}_{(3)}\right) \end{bmatrix},
\end{equation}
which is solved in a numerically more stable way using the pseudoinverse 
\begin{equation}
\text{vec}\left(\boldsymbol{G}^{+}\right) = \boldsymbol{K}^{\dagger}  \begin{bmatrix} \text{vec}\left(\boldsymbol{J}_{(3)}\right) \\ \vdots \\  \text{vec}\left(\boldsymbol{J}_{(3)}\right) \end{bmatrix}.
\end{equation}

In practice $\boldsymbol{K}$ has at least $r$ rank deficiencies. The underdetermined system results from the fact that $\boldsymbol{G}$ is estimated while shielded behind a finite difference. Every column $\boldsymbol{g}_i$ can therefore only be estimated up to an unknown constant term. The deficiencies can be removed by fixing one element of every column, e.g.\ setting it equal to zero. Note that no information on the constant terms can be obtained from first order derivative information (Jacobian tensor). The constant terms need to be estimated separately (Section \ref{s:DC}). 

In case a decomposition into $r > \text{rank}\left(\mathcal{J}\right)$, is required, the system of equations will remain underdetermined. In such cases the pseudoinverse ensures that the minimum norm solution is obtained. The $\text{rank}\left(\mathcal{J}\right)$ is here defined as the minimum number of terms for which the classical factorisation of \eqref{e:5} is exact.

\subsection{An explicit smoothness objective}
\label{a:B1}

\textbf{Update $\boldsymbol{G} \rightarrow \boldsymbol{G}^+$}

\begin{lem} Let \eqref{e:G} be the objective function from which $\boldsymbol{G}$ is to be updated. The claim is that $\boldsymbol{G}$ appears linearly in this joint objective such that an analytical update formula may be derived.
\begin{equation} \tag{\ref{e:G}}
\underset{\boldsymbol{G}}{\operatorname{arg~min}}~\left\| \boldsymbol{J}_{(3)} -(\mathcal{F}_C(\boldsymbol{V}) \circ \boldsymbol{G})(\boldsymbol{V} \kr \boldsymbol{W})^{\top} \right\|_F^2
+ \lambda \Vert \left(\mathcal{F}_L(\boldsymbol{V}) \circ \boldsymbol{G} \right) - \left(\mathcal{F}_R(\boldsymbol{V}) \circ \boldsymbol{G} \right) \Vert_F^2 
\end{equation}
\end{lem}

\begin{pf} Consider the vectorised form of the objective function. For clarity the dependency on $\boldsymbol{V}$ is dropped, e.g.\ $\mathcal{F}_C(\boldsymbol{V}) \coloneqq \mathcal{F}_C$.
\begin{equation} \label{e:obj_vec}
\underset{\boldsymbol{G}}{\operatorname{arg~min}}~\left\| \text{vec}(\boldsymbol{J}_{(3)}) -  \operatorname{vec}\left(\left(\mathcal{F}_C \circ \boldsymbol{G}\right)\left(\boldsymbol{V} \kr \boldsymbol{W} \right)^{\top} \right) \right\|_F^2
+\lambda \Vert  \text{vec}\left(\mathcal{F}_L \circ \boldsymbol{G} \right)  -\text{vec} \left(\mathcal{F}_R \circ \boldsymbol{G}\right) \Vert_F^2
\end{equation}
Also here we may invoke the property of \eqref{e:AXB}, together with \eqref{e:vec}. The objective is then rewritten into
\begin{equation}
\begin{split}
\underset{\boldsymbol{G}}{\operatorname{arg~min}}~&\left\| \text{vec}(\boldsymbol{J}_{(3)}) - \left( \left(\boldsymbol{V} \kr \boldsymbol{W}\right) \kron \boldsymbol{I}_N \right)\text{blkdiag}\left(\mathcal{F}_C\right) \text{vec}\left(\boldsymbol{G}\right) \right\|_F^2 \\
&+\lambda \Vert \left(\text{blkdiag}\left(\mathcal{F}_L\right) -\text{blkdiag} \left(\mathcal{F}_R \right) \right)\text{vec}\left(\boldsymbol{G}\right) \Vert_F^2.
\end{split}
\end{equation}
Reordering both terms leads to the expression
\begin{equation}
\underset{\boldsymbol{G}}{\operatorname{arg~min}}~\left\| \begin{bmatrix} \text{vec}(\boldsymbol{J}_{(3)}) \\ \boldsymbol{0} \end{bmatrix}
-  \begin{bmatrix}  \left( \left(\boldsymbol{V} \kr \boldsymbol{W}\right) \kron \boldsymbol{I}_N \right) \text{blkdiag}\left(\mathcal{F}_C \right) \\ \sqrt{\lambda}  \left(\text{blkdiag}\left(\mathcal{F}_L\right) -\text{blkdiag} \left(\mathcal{F}_R \right) \right) \end{bmatrix} \text{vec}(\boldsymbol{G}) \right\|_F^2,
\end{equation}
where $\boldsymbol{0} \in \mathbb{R}^{rN}$ is a vector of zeros and from which it is clear that $\text{vec}(\boldsymbol{G})$ enters linearly in the objective. \qed 
\end{pf}

An analytical update formula is then obtained by defining
\begin{equation}
\boldsymbol{K} \coloneqq  \begin{bmatrix}  \left( \left(\boldsymbol{V} \kr \boldsymbol{W}\right) \kron \boldsymbol{I}_N \right) \text{blkdiag}\left(\mathcal{F}_C \right) \\ \sqrt{\lambda}  \left(\text{blkdiag}\left(\mathcal{F}_L\right) -\text{blkdiag} \left(\mathcal{F}_R \right) \right) \end{bmatrix},
\end{equation}
leading to the least-squares solution
\begin{equation}
\text{vec}(\boldsymbol{G}^+) = \left(\boldsymbol{K}^{\top}\boldsymbol{K}\right)^{-1}\boldsymbol{K}^{\top} \begin{bmatrix} \text{vec}(\boldsymbol{J}_{(3)}) \\ \boldsymbol{0} \end{bmatrix},
\end{equation}
which is solved in a numerically more stable way using the pseudoinverse 
\begin{equation}
\text{vec}(\boldsymbol{G}^+) = \boldsymbol{K}^{\dagger} \begin{bmatrix} \text{vec}(\boldsymbol{J}_{(3)}) \\ \boldsymbol{0} \end{bmatrix}.
\end{equation}

To ensure that smoothness is promoted evenly on all columns of $\boldsymbol{G}$, a normalisation is required (similar to the update of $\boldsymbol{V}$). The block diagonal filters are therefore scaled by the root-mean-squared value of their corresponding columns. Denoting $\boldsymbol{G}'_L \coloneqq \mathcal{F}_L \circ \boldsymbol{G}$ we have that
\begin{equation}
\label{e:vec2}
\text{blkdiag}(\mathcal{F}_L) = \left[\begin{array}{cccc}\frac{\boldsymbol{F}_{L_1}}{\text{rms}(\boldsymbol{g}'_{L_1})} & 0 & \dots & 0 \\ 0 & \frac{\boldsymbol{F}_{L_2}}{\text{rms}(\boldsymbol{g}'_{L_2})} & \ddots & \vdots \\ \vdots & \ddots & \ddots & 0 \\ 0 & \cdots & 0 & \frac{\boldsymbol{F}_{L_r}}{\text{rms}(\boldsymbol{g}'_{L_r})} \end{array} \right].
\end{equation}
In an analogue way $\text{blkdiag}(\mathcal{F}_R)$ is normalised.

Also in this case $\boldsymbol{K}$ has at least $r$ rank deficiencies stemming from unknown constant terms. The deficiencies can be removed by fixing one element of every column, e.g.\ setting it equal to zero. The constant terms need to be estimated separately (Section \ref{s:DC}). Remaining deficiencies, encountered for decompositions where $r > \text{rank}\left(\mathcal{J}\right)$ are mitigated by the pseudoinverse, leading to the minimum norm solution.

\end{document}